# Diagnosis and Prognosis of COVID-19 Disease Using Routine Blood Values and LogNNet Neural Network

Mehmet Tahir Huyut [1,*] and Andrei Velichko [2,*]

[1] Department of Biostatistics and Medical Informatics, Faculty of Medicine, Erzincan Binali Yıldırım University, 24000 Erzincan, Türkiye
[2] Institute of Physics and Technology, Petrozavodsk State University, 33 Lenin Str., 185910 Petrozavodsk, Russia
* Correspondence: tahir.huyut@erzincan.edu.tr (M.T.H.); velichko@petrsu.ru (A.V.)

**Abstract:** Since February 2020, the world has been engaged in an intense struggle with the COVID-19 disease, and health systems have come under tragic pressure as the disease turned into a pandemic. The aim of this study is to obtain the most effective routine blood values (RBV) in the diagnosis and prognosis of COVID-19 using a backward feature elimination algorithm for the LogNNet reservoir neural network. The first dataset in the study consists of a total of 5296 patients with the same number of negative and positive COVID-19 tests. The LogNNet-model achieved the accuracy rate of 99.5% in the diagnosis of the disease with 46 features and the accuracy of 99.17% with only mean corpuscular hemoglobin concentration, mean corpuscular hemoglobin, and activated partial prothrombin time. The second dataset consists of a total of 3899 patients with a diagnosis of COVID-19 who were treated in hospital, of which 203 were severe patients and 3696 were mild patients. The model reached the accuracy rate of 94.4% in determining the prognosis of the disease with 48 features and the accuracy of 82.7% with only erythrocyte sedimentation rate, neutrophil count, and C reactive protein features. Our method will reduce the negative pressures on the health sector and help doctors to understand the pathogenesis of COVID-19 using the key features. The method is promising to create mobile health monitoring systems in the Internet of Things.

## 1. Introduction

The new severe acute respiratory syndrome coronavirus (SARS-CoV-2), first identified in 2019, has rapidly affected the world and caused a pandemic [1,2]. The disease, identified as coronavirus 2019 (COVID-19), can cause severe pneumonia and fatal acute respiratory distress syndrome (ARDS) [3–6]. While the disease may be asymptomatic, severe ARDS is thought to be caused by an inflammatory cytokine storm that may be encountered during the disease period [6,7]. The pathogen can cause a serious respiratory disorder that requires special intervention in intensive care units (ICUs) and, in some cases, may cause death [6,7]. Moreover, the symptoms of COVID-19 induced by the new SARS-CoV-2 are difficult to distinguish from known infections in the majority of patients [6,8,9].

Previous studies have demonstrated the clinical importance of changes in routine blood parameters (RBV) in the diagnosis and prediction of prognosis of infectious diseases [1–4,10–12]. Similarly, many abnormalities have been reported in the peripheral blood of patients infected with COVID-19 [6,7,11]. However, Jiang et al. [13] and Zheng et al. [14] emphasized that information on early predictive factors for particularly severe and fatal COVID-19 cases is relatively limited and further research is needed. Huyut et al. [6] and Lippi et al. [15] described that the rapid spread of disease in pandemics overwhelms health systems and raises concerns about the need for intensive care treatment [6,15]. In addition, the detection of severe and mild patients in COVID-19 is an important and clinically difficult process in terms of morbidity and mortality [6]. Despite these clinical features of COVID-19, studies with large samples representing laboratory abnormalities of patients are needed [3,16]. Therefore, the relationship between COVID-19 disease and RBVs should be supported by large datasets.

Studies have sought how to determine whether patients who are likely to benefit from supportive care and early intervention are at risk and how to identify them [6,11]. While new tests are being developed for the diagnosis of COVID-19, Banerjee et al. [8] stated that these applications require specialized equipment and facilities. Estimating the diagnosis and prognosis of diseases without using advanced devices and methods can help with various problems, such as patient comfort, as well as health system and economic inefficiencies. For this purpose, Beck et al. [17] and Xu et al. [18] have reported that more economical and faster alternative methods are being developed to assist clinical procedures.

Uncertainties in the routine blood values of COVID-19 patients, in addition to difficulties in diagnosis and treatment have increased the interest in machine learning (ML) and artificial intelligence (AI) approaches. Artificial intelligence models have the power to reveal hidden relationship structures between features [19]. Artificial intelligence approaches are frequently used in real-time decision making to reduce drug costs, improve patient comfort, and improve the quality of healthcare services [5,19].

There are several artificial intelligence methods to predict the diagnosis and mortality of COVID-19 [4,17]. Most of these studies have relied on computed tomography (CT) [19], while far fewer studies relied on RBVs [4,5,20]. Imaging-based solutions are costly, time-consuming, and require specialized equipment [20]. Diagnosis based on RBV values can provide an effective, rapid, and cost-effective alternative for the early detection and prognosis of COVID-19 cases [5,20,21].

Previous AI studies did not use most of the RBV parameters and reported relatively poor classifier performance compared to the current study [2,3,5,6]. In addition, previous studies [8,19–25] have generally focused on the early diagnosis of COVID-19 disease and have addressed relatively smaller samples. Artificial intelligence studies on predicting the prognosis of the disease and detecting severely or mildly infected patients in the early period based on RBVs alone are insufficient. New studies could reduce the intensity of the ICU and help health services by detecting severe and mildly infected patients with COVID-19 early [2,5,19,20].

Most ML approaches involve the process of transforming the feature vector from the first multidimensional space to the second multidimensional space and detecting the vector by a linear classifier [26]. The differences between ML models generally lie in the transformation algorithms and their number and order. In addition, transformation algorithms can be in the form of reducing and increasing the space dimension. The popular machine learning classifier algorithms used for data analysis are: multilayer perceptron (feedforward neural network with several layers, linear classifier) [27], support vector machine [28], K-nearest neighbors method [29], XGBoost classifier [30], random forest method [31], logistic regression [32], and decision trees [33].

ML algorithms typically require a sufficiently large number of samples. However, in our case, the dataset has to be reduced to avoid dimensionality problems by finding a matrix that has fewer columns and is similar to the original matrix. Since the new matrix consists of fewer features, it can be used more efficiently than the original matrix. Dimensionality reduction is the process of finding matrices with fewer columns. Feature selection is one of the techniques used to reduce dimensionality, when irrelevant and redundant features are discarded [26,34]. In addition, the selection of appropriate features can reduce the measurement cost and provide a better understanding of the problem [26]. Feature selection methods can be classified as filters, embedded methods, and wrappers (forward selection, backward elimination, recursive feature elimination) [26,34]. Because feature selection is part of the training process in embedded methods, our method lies between filters and wrappers. Searching for the best subset of features is performed during training of the classifier, e.g., when optimizing weights in a neural network. Therefore, embedded methods present a lower computational cost than wrappers [26].

Most of the feature selection methods are filters, although we can find representative methods for all three categories [26]. The large number of available feature selection methods complicates the selection of the best method for a given problem [34]. The latest methods that have become popular among researchers are feature selection based on correlation (CFS) [35], filtering based on consistency [36], INTERACT [37], knowledge gain (InfoGain) [38], ReliefF [39], recursive feature elimination for support vector machines (SVM-RFE) [40], Lasso editing [41], and the minimum redundancy maximum relevance (mRMR) algorithm (developed specifically for dealing with microarray data) [26].

In [42], a classifier based on the LogNNet neural network was described using a handwriting recognition example from the MNIST database. Velichko [43] demonstrated the use of the LogNNet to calculate risk factors for the presence of a disease based on a set of medical health indicators. The LogNNet neural network is a feedforward network that improves classification accuracy by passing the feature vector through a special reservoir matrix and transforming it into a feature vector of different size [44]. Previous studies have shown that the higher the entropy of a chaotic mapping that fills a reservoir matrix, the better the classification accuracy [45]. Therefore, the procedure for optimizing chaotic map parameters plays an important role in the presented data analysis method using the LogNNet neural network. In addition, due to the characteristics of chaotic mapping, RAM usage by a neural network can be significantly reduced. In [43], the operation of the LogNNet algorithm on a device with 2 kB of RAM was presented. This result demonstrated that LogNNet can be used in Internet of Things (IoT) mobile devices.

In this study, we apply the LogNNet neural network for the diagnosis and prognosis of COVID-19 using the RBV values measured at the time of admission to the hospital. The wrapper-type backward feature elimination algorithm has been successfully adapted to LogNNet. The novelty of the presented method is the approach to the diagnosis and prognosis of COVID-19 using routine blood values.

The paper has the following structure. Section 2 describes the data collection procedure, the basic LogNNet architecture, and K-fold cross-validation technique. Section 3 presents examples of using the feature selection methodology for two datasets. In this section, the most important RBVs (features) effective in the diagnosis and prognosis of the disease were selected. Using various feature combinations, the performance of the LogNNet model in the diagnosis and prognosis of the disease was calculated. Section 4 discusses the results and compares them with known developments. In conclusion, a general description of the study and its scientific significance are given.

## 2. Materials and Methods

This study was conducted in accordance with the Declaration of Helsinki, 1989. Data were collected retrospectively from the information system of Erzincan Binali Yıldırım University Mengücek Gazi Training and Research Hospital (EBYU-MG) between April and December 2021. The study had three main stages: data collection, LogNNet training with selection of main features, and testing of feature combinations (Figure 1).

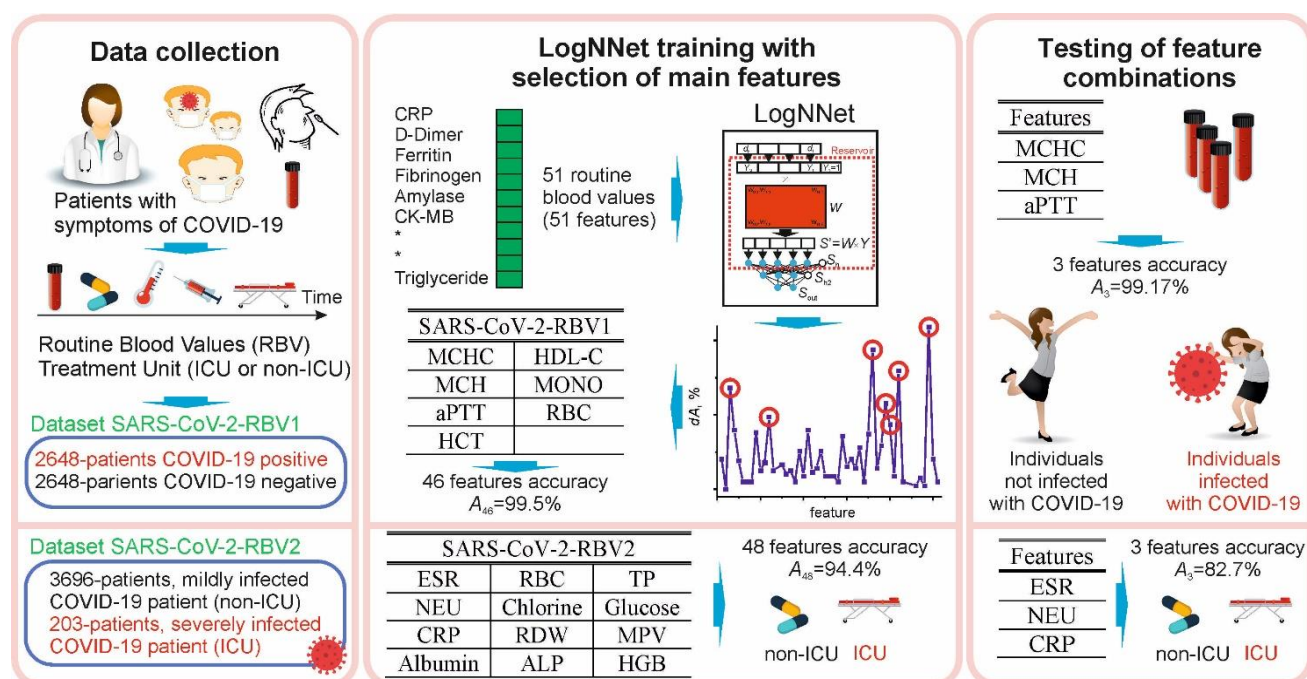

**Figure 1.** The main stages of the study for the diagnosis and prognosis of COVID-19 using the routine blood values: data collection, LogNNet training with the selection of main features, testing combinations of the most important features that influence the diagnosis and prognosis of the disease.

The RBV of the patients consisted of biochemical, hematological, and immunological tests. Patients admitted to the ICU were defined as severely infected, while patients who could not be admitted to the ICU (non-ICU, subjects in all wards) were defined as mildly infected. The dataset SARS-CoV-2-RBV1 included information on $n$ = 2648 COVID-19 positive outpatients and $n$ = 2648 COVID-19 negative (control group), for a total of 5296 patients. The dataset SARS-CoV-2-RBV2 contained information of $n$ = 203 ICU and $n$ = 3696 non-ICU COVID-19 patients. Raw data records included patients' diagnoses (COVID-19, heart disease, asthma, etc.), treatment units (ICU or non-ICU), age, and RBV data. The entire recording process took 20 h. In the raw data, RBV data were on a quantitative scale, diagnostic data were on a multinomial scale, and treatment units were on a binomial scale. In the data preprocessing stage, the string data were converted into numerical data. Categorical data were coded, repeated measurements were averaged, duplicates were removed, and quantitative data were normalized. The missing RBV data were complemented by the mean of the respective parameter distribution.

## 2.1. Characteristic of Participants, Workflow and Define Datasets

In the EBYU-MG hospital, only the cases that were detected as SARS-CoV-2 by real-time reverse transcriptase polymerase chain reaction (RT-PCR) in nasopharyngeal or oropharyngeal swabs during the dates covered by this study were diagnosed with COVID-19. The research only included individuals over the age of 18. In order to prevent various complications, RBV results at the first admission were recorded.

The first SARS-CoV-2-RBV dataset (SARS-CoV-2-RBV1) includes the information of 2648 patients diagnosed with COVID-19 and receiving outpatient treatment in hospital on the specified dates, and the same number of patients (control group) whose COVID-19 tests were negative. The control group was randomly selected from individuals over the age of 18 who had applied to the emergency COVID-19 service but had a negative RT-PCR test. With the feature selection procedure, the most important RBV features that are effective in the diagnosis of the disease were selected from the SARS-CoV-2-RBV1 dataset.

The selected features were fed into LogNNet neural network to examine the method's performance in diagnosing COVID-19 disease.

The second SARS-CoV-2-RBV dataset (SARS-CoV-2-RBV2) includes the information of 3899 patients who were treated for COVID-19 in hospital on the specified dates. The treatment units of these patients at the first admission were examined. The SARS-CoV-2-RBV2 dataset contains $n$ = 203 ICU and $n$ = 3696 non-ICU COVID-19 patients. Then, with the feature selection procedure, the most influential RBV traits in the prognosis of the disease were selected from the SARS-CoV-2-RBV2 dataset. Selected features were fed into the LogNNet neural network to examine the performance of this method in determining the prognosis and severity of COVID-19 disease.

The SARS-CoV-2-RBV1 and SARS-CoV-2-RBV2 datasets are presented in Tables 1 and 2. SARS-CoV-2-RBV1 and SARS-CoV-2-RBV2 datasets include immunological, hematological, and biochemical RBV parameters and each dataset consists of 51 features. In the SARS-CoV-2-RBV1 dataset, positive COVID-19 test results were coded as 1 and negative as 0 (COVID-19 = 1, non-COVID-19 = 0).

In the SARS-CoV-2-RBV2 dataset, severely infected (ICU) COVID-19 patients were coded as 1, while mildly infected (non-ICU) COVID-19 patients were coded as 0. Datasets are available for download in the Supplementary Materials.

**Table 1.** Feature numbering for SARS-CoV-2-RBV1 datasets.

| № | Feature | № | Feature | № | Feature | № | Feature | № | Feature |
|---|---|---|---|---|---|---|---|---|---|
| 1 | CRP | 12 | NEU | 23 | MPV | 34 | GGT | 45 | Sodium |
| 2 | D-Dimer | 13 | PLT | 24 | PDW | 35 | Glucose | 46 | T-Bil |
| 3 | Ferritin | 14 | WBC | 25 | RBC | 36 | HDL-C | 47 | TP |
| 4 | Fibrinogen | 15 | BASO | 26 | RDW | 37 | Calcium | 48 | Triglyceride |
| 5 | INR | 16 | EOS | 27 | ALT | 38 | Chlorine | 49 | eGFR |
| 6 | PT | 17 | HCT | 28 | AST | 39 | Cholesterol | 50 | Urea |
| 7 | PCT | 18 | HGB | 29 | Albumin | 40 | Creatinine | 51 | UA |
| 8 | ESR | 19 | MCH | 30 | ALP | 41 | CK | | |
| 9 | Troponin | 20 | MCHC | 31 | Amylase | 42 | LDH | | |
| 10 | aPTT | 21 | MCV | 32 | CK-MB | 43 | LDL | | |
| 11 | LYM | 22 | MONO | 33 | D-Bil | 44 | Potassium | | |

CRP: C-reactive protein; INR: international normalized ratio; PT: prothrombin time; PCT: Procalcitonin; ESR: erythrocyte sedimentation rate; aPTT: activated partial prothrombin time; LYM: lymphocyte count; NEU: neutrophil count; PLT: platelet count; WBC: white blood cell count; BASO: basophil count; EOS: eosinophil count; HCT: hematocrit; HGB: hemoglobin; MCH: mean corpuscular hemoglobin; MCHC: mean corpuscular hemoglobin concentration; MCV: mean corpuscular volume; MONO: monocyte count; MPV: mean platelet volume; PDW: platelet distribution width; RBC: red blood cells; RDW: red cell distribution width; ALT: alanine aminotransaminase; AST: aspartate aminotransferase; ALP: alkaline phosphatase; CK-MB: creatine kinase myocardial band; D-Bil: direct bilirubin; GGT: gamma-glutamyl transferase; HDL-C: high-density lipoprotein-cholesterol; CK: creatine kinase; LDH: lactate dehydrogenase; LDL: low-density lipoprotein; T-Bil: total bilirubin; TP: total protein; eGFR: estimating glomerular filtration rate; UA: uric acid.

**Table 2.** Feature numbering for SARS-CoV-2-RBV2 datasets.

| № | Feature | № | Feature | № | Feature | № | Feature | № | Feature |
|---|---|---|---|---|---|---|---|---|---|
| 1 | ALT | 12 | Chlorine | 23 | eGFR | 34 | MONO | 45 | Fibrinogen |
| 2 | AST | 13 | Cholesterol | 24 | Urea | 35 | MPV | 46 | INR |
| 3 | Albumin | 14 | Creatinine | 25 | UA | 36 | NEU | 47 | PT |
| 4 | ALP | 15 | CK | 26 | BASO | 37 | PDW | 48 | PCT |
| 5 | Amylase | 16 | LDH | 27 | EOS | 38 | PLT | 49 | ESR |
| 6 | CK-MB | 17 | LDL | 28 | HCT | 39 | RBC | 50 | Troponin |
| 7 | D-Bil | 18 | Potassium | 29 | HGB | 40 | RDW | 51 | aPTT |
| 8 | GGT | 19 | Sodium | 30 | LYM | 41 | WBC | | |
| 9 | Glucose | 20 | T-Bil | 31 | MCH | 42 | CRP | | |
| 10 | HDL-C | 21 | TP | 32 | MCHC | 43 | D-Dimer | | |
| 11 | Calcium | 22 | Triglyceride | 33 | MCV | 44 | Ferritin | | |

ALT: alanine aminotransaminase; AST: aspartate aminotransferase; ALP: alkaline phosphatase; CK-MB: creatine kinase myocardial band; D-Bil: direct bilirubin; GGT: gamma-glutamyl transferase; HDL-C: high-density lipoprotein-cholesterol; CK: creatine kinase; LDH: lactate dehydrogenase; LDL: low-density lipoprotein; T-Bil: total bilirubin; TP: total protein; eGFR: estimating glomerular filtration rate; UA: uric acid; BASO: basophil count; EOS: eosinophil count; HCT: hematocrit; HGB: hemoglobin; LYM: lymphocyte count; MCH: mean corpuscular hemoglobin; MCHC: mean corpuscular hemoglobin concentration; MCV: mean corpuscular volume; MONO: monocyte count; MPV: mean platelet volume; NEU: neutrophil count; PDW: platelet distribution width; PLT: platelet count; RBC: red blood cells; RDW: red cell distribution width; WBC: white blood cell count; CRP: C-reactive protein; INR: international normalized ratio; PT: prothrombin time; PCT: procalcitonin; ESR: erythrocyte sedimentation rate; aPTT: activated partial prothrombin time.

*2.2. LogNNet Architecture*

Figure 2 demonstrates the principle of operation of the neural network LogNNet [43].

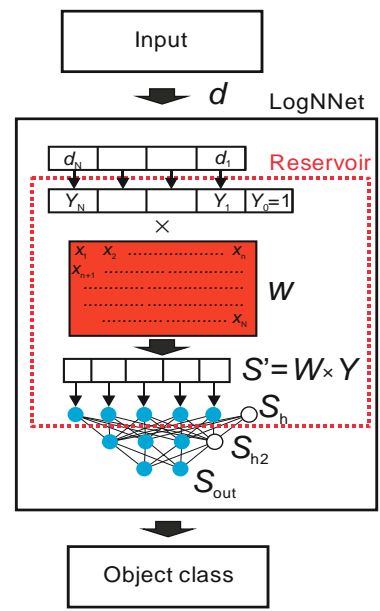

**Figure 2.** LogNNet architecture [43].

An object in the form of a feature vector, denoted as *d*, is inputted to LogNNet. The feature vector contains *N* coordinates ($d_1, d_2 \ldots d_N$), where the number *N* is defined by the user. The classifier output determines the object class to which the input feature vector *d* belongs. The number of possible classes is denoted as *M*. LogNNet contains a reservoir with a special matrix, denoted as *W*. The matrix *W* was filled in a row-by-row pattern with numbers generated by the chaotic mapping $x_n$. We use chaotic mapping based on the congruential generator Equation (1) (see Table 3) and the algorithm of matrix *W* filling shown in Algorithm 1. Vector *d* is converted into a vector *Y* of dimension *N* + 1 with an additional coordinate $Y_0$ = 1, and each component is normalized by dividing by the maximum value of this component in the training base. The next step is a multiplication of a special matrix *W* with the dimension (*N* + 1) × *P* and a vector *Y*. The result is a vector *S'* with *P* coordinates, which is normalized [42] and converted into a vector $S_h$ of dimension *P + 1* with zero coordinate $S_h$[0] = 1, which plays the role of a bias element. In this way, the primary transformation of the feature vector *d* into the second *(P + 1)*-dimensional space is completed. Then, the vector $S_h$ is fed to a two-layer linear classifier, with the number of neurons *H* in the hidden layer $S_{h2}$, and the number of outputs *M* in the output layer $S_{out}$. To indicate the parameters of the neural network, the following designation LogNNet *N:P:H:M* is used.

**Table 3.** Chaotic map equation and list of optimized parameters with limits.

| Chaotic Map | List of Optimized Parameters (Limits) | Equation | |
|---|---|---|---|
| Congruent generator | $K$ (−100 to 100) $D$ (−100 to 100) $L$ (2 to 10,000) $C$ (−100 to 100) | $\begin{cases} x_{n+1} = (D - K \cdot x_n) \mod L \\ x_1 = C \end{cases}$ | (1) |

**Algorithm 1.** Algorithm of matrix *W* filling.

```
xn: = C;
for j: = 1 to P do
for i: = 0 to N do
begin
xn: = (D-K * xn) mod L; // Congruential generator formula
W [i,j]: = xn/L;
end;
```

The training of the linear classifier LogNNet was carried out using the backpropagation method [42].

*2.3. Optimization of Reservoir Parameters*

The optimal chaotic mapping parameters were selected using a special algorithm. The ranges of the parameters are indicated in Table 3. Before optimization, it is necessary to set the following values of the constant parameters of the model: the value *P* + 1, which determines the dimension of the vectors $S_h$ and $S_{h2}$, the number of layers in the linear classifier, the number of epochs *Ep* for backpropagation training, and the number of neurons in the classifier's hidden layer, in the case of a two-layer classifier. The training of the LogNNet network is performed by two nested iterations [46]. The inner iteration trains the output LogNNet classifier by backpropagation of error on the training set, and the outer iteration optimizes the model parameters.

During the optimization process, the training and validation bases coincided and were equivalent to the initial datasets (SARS-CoV-2-RBV1 or SARS-CoV-2-RBV2). The outer iteration implements the particle swarm method with fitness function equal to classification accuracy. Outer iteration ends either when the desired values of the classification accuracy are reached, or when the specified number of iterations in the particle swarm method is completed. As a result, the optimized model parameters (chaotic mapping parameters) at the output allow us to obtain the highest classification accuracy on the validation set.

*2.4. Classification Accuracy, K-Fold Cross-Validation and Balancing Techniques*

The K-fold cross-validation technique was used to test LogNNet. This method is well suited for the medical databases, which are not split into test and training sets. The elements of the set (SARS-CoV-2-RBV1 or SARS-CoV-2-RBV2) are divided into *K* parts (*K* = 5). One of the parts is taken as the test sample, and the remaining *K*-1 parts are used for the training sample. Then, the average value of the metrics is calculated for all *K* cases when one of the *K* parts of the set becomes the test sample in turn. A distinctive feature of the method is that the separate test data are not needed for the training process. Applying the K-fold cross-validation technique, we calculate the classification metrics: classification accuracy, *A*, precision, recall, and F1-metric. Wherever we talk about the classification accuracy *A* in this article, we imply the value obtained by the K-fold cross-validation method.

To obtain a higher value of *A*, the training *K*-1 parts of the sets were balanced as in [43]. The balancing implies equalizing the number of objects for each class, supplementing the classes with copies of already existing objects, and sorting the training set in sequential order. The balancing process can be illustrated by the following example. The training set consists of 10 objects divided into 2 classes. Each object is assigned a feature vector $d z_m$, where *z* is the object number *z* = 1, …, 10, *m* is the class number *m* = 1…2. For example, we have 7 objects of class 1 ($d1_1$, $d2_1$, $d4_1$, $d5_1$, $d6_1$, $d7_1$, $d10_1$) and three objects of class 2 ($d3_2$, $d8_2$, $d9_2$). We find the maximum number of objects (*MAX*) in the classes, and *MAX* equals 7 for class 1. We supplement the remaining groups with copies of the already existing objects (duplication) to equalize the number to *MAX*. Therefore, for class 2, we acquire the group ($d3_2$, $d8_2$, $d9_2$, $d3_2$, $d8_2$, $d9_2$, $d3_2$). Then, we compose a balanced training data set, choosing one object from each group in turn. As a result, we achieve the following training set: ($d1_1$, $d3_2$, $d2_1$, $d8_2$, $d4_1$, $d9_2$, $d5_1$, $d3_2$, $d6_1$, $d8_2$, $d7_1$, $d9_2$, $d10_1$, $d3_2$), which consists of 14 vectors and has the same number of objects in every class.

*2.5. Threshold Approach*

The simplest approach for classifying by one feature in the presence of only two classes is based on determining the threshold value separating the classes *Vth*. For the SARS-CoV-2-RBV1 dataset, we introduce an additional designation of the type of threshold value Type 1 or Type 2 in accordance with the rule:

$$\begin{cases} \text{Type 1:} & \text{if feature value} > Vth \text{ then "Covid-19" else "non-Covid-19"} \\ \text{Type 2:} & \text{if feature value} > Vth \text{ then "non-Covid-19" else "Covid-19"} \end{cases} \quad (2)$$

The threshold type indicates which side of the threshold the sick and healthy classes are on.

For the SARS-CoV-2-RBV2 dataset (after balancing, see Section 2.4), we introduce a similar relation for the type of threshold value:

$$\begin{cases} \text{Type 1:} & \text{if feature value} > Vth \text{ then "ICU" else "non-ICU"} \\ \text{Type 2:} & \text{if feature value} > Vth \text{ then "non-ICU" else "ICU"} \end{cases} \quad (3)$$

Threshold accuracy after balancing datasets (see Section 2.4) is determined as

$$Ath = \frac{TP + TN}{TP + TN + FP + FN} \quad (4)$$

were *TP* denotes true positive, *TN* true negative, *FP* false positive, and *FN* false negative.

K-fold validation is not used when calculating *Ath*.

The threshold value *Vth* was determined by stepwise enumeration and finding the maximum value of *Ath*.

The threshold method reflects the dependence of one feature and COVID-19 and indicates the classification success (Equations (2)–(4)). In practical applications, the LogNNet is a more powerful classification tool than the simple threshold method, revealing more information between features and COVID-19.

*2.6. Feature Selection Method*

The feature selection method is based on a wrapper-type backward feature elimination algorithm and has two consecutive steps. First, redundant features and features that make training of the neural network difficult are removed. In backward elimination, the algorithm starts with all the features and removes the least significant feature at each iteration. The features are removed by zeroing the corresponding components of the input vectors *d*. The second stage includes sorting the remaining features according to their contribution to the classification metric.

Features selection for the dataset SARS-CoV-2-RBV2 illustrates this method. Let us suppose a reservoir optimization was carried out and an accuracy of $A_{51}$ = 93.665% was obtained (using K-fold cross-validation), where the designation $A_{NF}$ means the classification accuracy when using *NF* = 51 features. Let us introduce additional pointers, denote the set of removed features by *FR*, and denote the set of selected features by *FS*. For example, $A_{49}$(*FR* [3,33]) denotes accuracy at *NF* = 49 features with features *z* = 3 and *z* = 33 removed, and $A_4$(*FS* [1,22,33,41,55] denotes accuracy at *NF* = 4 features with the main features from the set *FS*, *z* = 1, 22, 33, 41, 55. Next, we plot the dependence of the value of $dA_{51}$ on the number of the removed feature *z* (see Figure 3a), where

$$dA_{51}(z) = A_{51} - A_{50}(FR[z]) \tag{5}$$

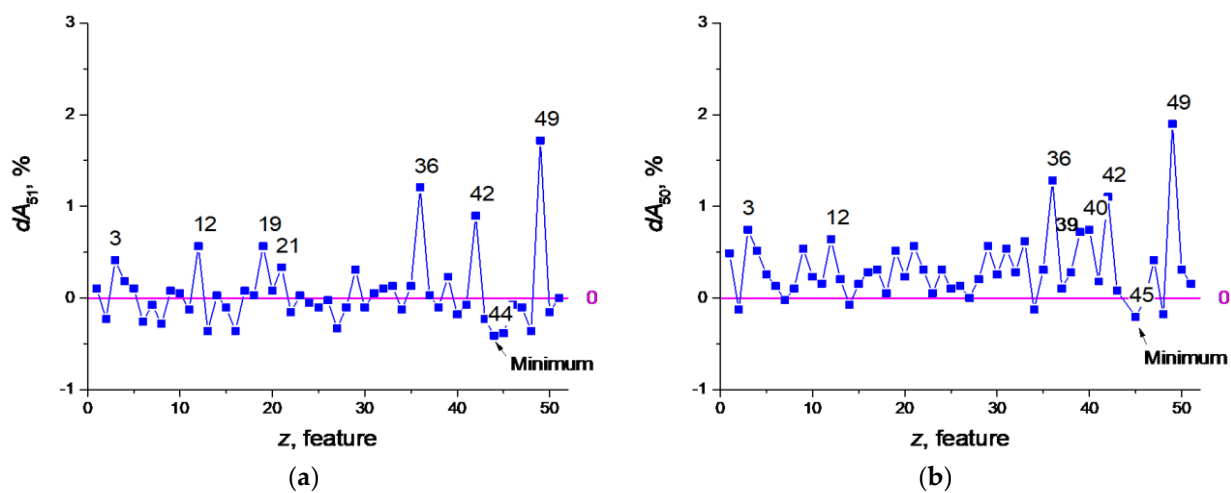

(a) (b)

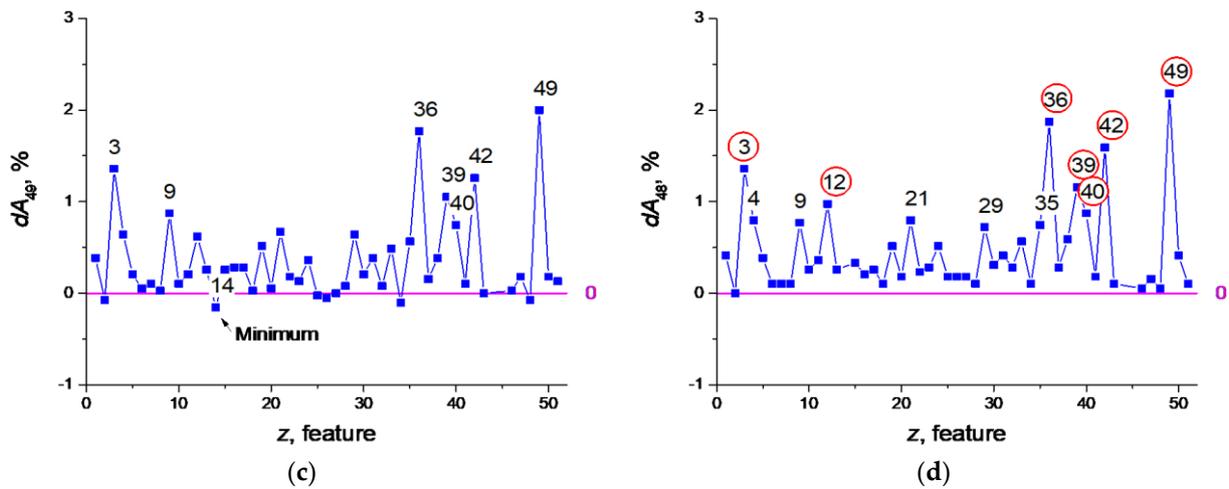

**Figure 3.** Function of the feature strength $dA_{51}(z)$ (**a**), $dA_{50}(z)$ (**b**), $dA_{49}(z)$ (**c**), $dA_{48}(z)$ (**d**).

Dependence $dA(z)$ is a function of the feature strength. The value $A_{50}(FR[z])$ characterizes the classification accuracy of the neural network using $NF = 50$ features, after deleting the feature with number $z$. Positive feature strength $dA_{51}$ (Figure 3a and Equation (5)) means that the removal of the feature reduces the classification accuracy of the network and the feature is useful. Negative $dA_{51}$ means that the feature interferes with learning (redundant) and its removal leads to an increase in the classification properties of the neural network. After the first selection iteration, the seven most useful features can be identified having numbers $z = 49, 36, 42, 19, 12, 3, 21$ (Figure 3a). The feature that makes learning the most difficult is number $z = 44$ (in Figure 3 it is indicated by the index 'Minimum'). Its removal makes $A_{50}(FR[44]) = 94.075\%$, which exceeds the previous value $A_{51} = 93.665\%$.

The next iteration involves calculating the dependence of $dA_{50}(z)$ (Figure 3b), where

$$dA_{50}(z) = A_{50}(FR[44]) - A_{49}(FR[44, z]) \qquad (6)$$

Equation (6) implies the exclusion of the worst feature $z = 44$ and the exclusion of all other features in turn. As a result, the next feature to exclude will be the feature $z = 45$, and the best accuracy will be $A_{49}(FR[44,45]) = 94.28\%$.

Iterations continue until all $dA$ values are greater than or equal to zero. Figure 3c,d shows graphs for Equations (7) and (8)

$$dA_{49}(z) = A_{49}(FR[44, 45]) - A_{48}(FR[44, 45, z]) \qquad (7)$$

$$dA_{48}(z) = A_{48}(FR[44, 45, 14]) - A_{47}(FR[44, 45, 14, z]) \qquad (8)$$

The graph in Figure 3d reflects the dependence $dA_{48}(z)$ that has positive values. Thus, the best classification accuracy corresponds to $A_{48}(FR[14,44,45]) = 94.434\%$, after removing the features $z = 44, 45, 14$. During the selection, the set of the seven best features with highest feature strength $dA$ also changed from the set [3,12,19,21,36,42,49] (Figure 3a) to [3,12,36,39,40,42,49] (Figure 3d, red circle).

The second stage arranges the features according to their strength in descending order of peak values $dA$. For the considered example, the sequence contains the following first 12 values [3,4,9,12,21,29,35,36,39,40,42,49] (Figure 3d).

## 3. Results

### 3.1. Dataset SARS-CoV-2-RBV1

LogNNet 51:50:20:2 architecture was used for SARS-CoV-2-RBV1 dataset. Reservoir optimization following the method from Section 2.3 with the number of epochs $Ep = 50$ led to the parameters of the congruential generator listed in Table 4.

**Table 4.** Optimal reservoir parameters.

| Dataset SARS-CoV-2-RBV1 | | | | Dataset SARS-CoV-2-RBV2 | | | |
|---|---|---|---|---|---|---|---|
| *K* | *D* | *L* | *C* | *K* | *D* | *L* | *C* |
| 93 | 68 | 9276 | 73 | 47 | 99 | 8941 | 56 |

Feature selection was performed with the number of epochs $Ep = 100$. Prior to selection, the $dA_{51}(z)$ shape is plotted in Figure 4a. After feature selection, the redundant features have the numbers $z = 21, 37, 42, 49, 40$, and the $dA_{46}(z)$ plot is shown in Figure 4b. The influence of features with numbers $z = 20, 19, 10, 17$ has increased.

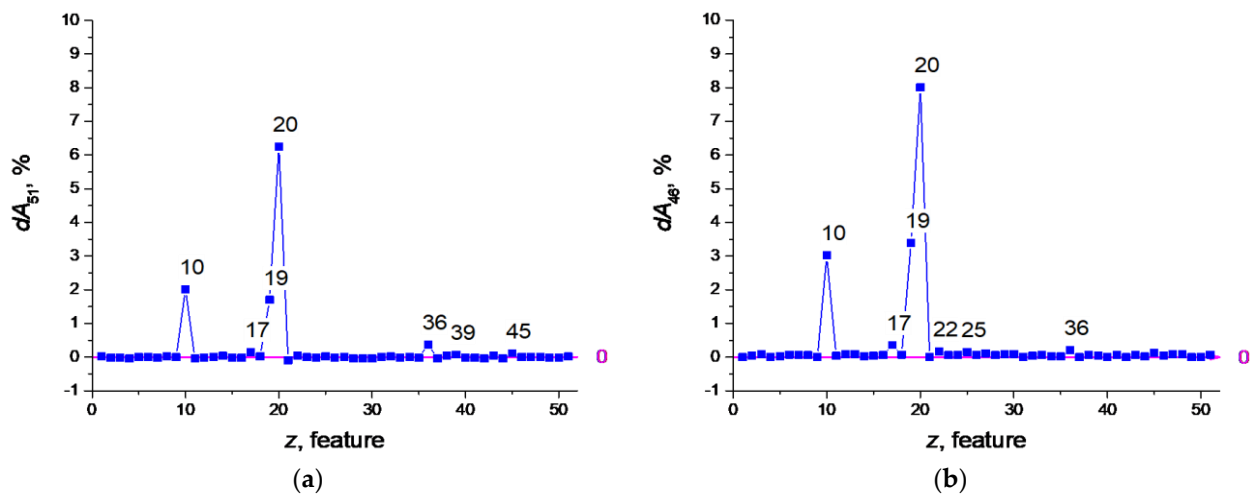

(a)　　　(b)

**Figure 4.** Function of the feature strength $dA_{51}(z)$ (**a**), $dA_{46}(z)$ (**b**).

The dependence of $A_{46}(FR\,[21,37,40,42,49])$ on the number of epochs is shown in Figure 5, and the values of other metrics are shown in Table 5.

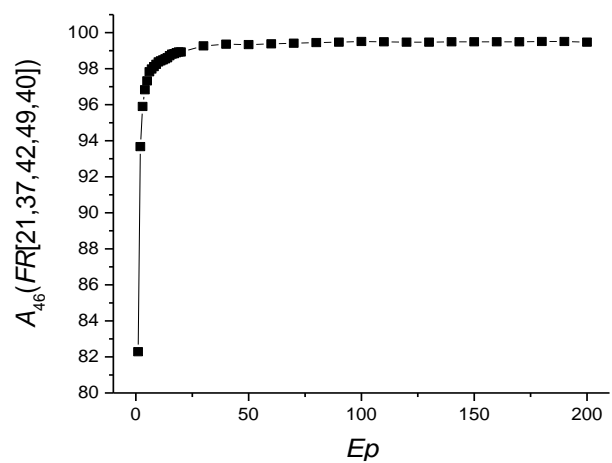

**Figure 5.** Dependence of $A_{46}(FR\,[21,37,40,42,49])$ on the number of epochs $Ep$.

Table 5. Classification metrics depending on the number of training epochs *Ep*.

| Ep | $A_{46}$(FR [21,37,40,42,49]) | Precision "Non-COVID-19" | Precision "COVID-19" | Recall "Non-COVID-19" | Recall "COVID-19" | F1 "Non-COVID-19" | F1 "COVID-19" |
|---|---|---|---|---|---|---|---|
| 10 | 98.376 | 0.978 | 0.99 | 0.991 | 0.977 | 0.984 | 0.984 |
| 30 | 99.339 | 0.992 | 0.995 | 0.995 | 0.992 | 0.993 | 0.993 |
| 100 | 99.509 | 0.994 | 0.996 | 0.996 | 0.994 | 0.995 | 0.995 |
| 150 | 99.49 | 0.994 | 0.996 | 0.996 | 0.994 | 0.995 | 0.995 |
| 200 | 99.471 | 0.994 | 0.995 | 0.995 | 0.994 | 0.995 | 0.995 |

*Ep* = 100 will be taken as the optimal value of the number of epochs. The RBV values found most important in the diagnosis of COVID-19 are the features listed in Table 6. The most important of these are MCHC, MCH, and aPTT. MCHC in a blood test allows to find out the average amount of hemoglobin in an erythrocyte.

Table 6. The seven features found to be most important in the diagnosis of COVID-19.

| Number | $dA_{46}$ | Features |
|---|---|---|
| 20 | 8.007 | MCHC |
| 19 | 3.399 | MCH |
| 10 | 3.022 | aPTT |
| 17 | 0.359 | HCT |
| 36 | 0.208 | HDL-C |
| 22 | 0.17 | MONO |
| 25 | 0.151 | RBC |

MCH: corpuscular hemoglobin; MCHC: corpuscular hemoglobin concentration; aPTT: activated partial prothrombin time; HCT: hematocrit; HDL-C: high-density lipoprotein-cholesterol; MONO: monocyte count; RBC: red blood cells.

The efficiency of LogNNet in determining the diagnosis of COVID-19 using only seven features and their combinations is shown in Table 7.

Table 7. LogNNet efficiency for various combinations of features.

| Combinations of Features | A | Precision "Non-COVID-19" | Precision "COVID-19" | Recall "Non-COVID-19" | Recall "COVID-19" | F1 "Non-COVID-19" | F1 "COVID-19" |
|---|---|---|---|---|---|---|---|
| $A_{46}$(FR [21,37,40,42,49]) | 99.509 | 0.994 | 0.996 | 0.996 | 0.994 | 0.995 | 0.995 |
| $A_7$(FS [10,17,19,20,22,25,36]) | 99.358 | 0.991 | 0.996 | 0.996 | 0.991 | 0.994 | 0.994 |
| $A_1$(FS [20]) | 94.279 | 0.930 | 0.958 | 0.959 | 0.926 | 0.944 | 0.942 |
| $A_1$(FS [19]) | 52.418 | 0.526 | 0.524 | 0.500 | 0.548 | 0.509 | 0.532 |
| $A_1$(FS [10]) | 52.398 | 0.516 | 0.947 | 0.972 | 0.075 | 0.672 | 0.100 |
| $A_1$(FS [36]) | 94.429 | 0.935 | 0.955 | 0.956 | 0.932 | 0.945 | 0.943 |
| $A_2$(FS [19,20]) | 99.150 | 0.989 | 0.994 | 0.994 | 0.989 | 0.992 | 0.991 |
| $A_2$(FS [20,36]) | 97.583 | 0.973 | 0.979 | 0.979 | 0.972 | 0.976 | 0.976 |
| $A_2$(FS [19,36]) | 94.373 | 0.934 | 0.955 | 0.957 | 0.931 | 0.945 | 0.943 |
| $A_3$(FS [10,19,20]) | 99.169 | 0.989 | 0.995 | 0.995 | 0.989 | 0.992 | 0.992 |
| $A_5$(FS [10,17,19,22,25]) | 51.699 | 0.526 | 0.546 | 0.784 | 0.250 | 0.604 | 0.277 |

Using only one feature 20 (MCHC) or 36 (HDL-C) in determining the diagnosis of COVID-19 provides a high classification accuracy of $A_1$(FS [20]), $A_1$(FS [36]) ~94%. The combination of 2 features 20 (MCHC) and 19 (MCH) allows to reach accuracy $A_2(FS$ [19,20]) ~99.15%.

The accuracy of the model in diagnosing the disease with seven features was almost equal to the accuracy rate in using all 46 features ($A_7$~99.4 vs. $A_{46}$~99.59) (Table 7).

Threshold Accuracy on One Feature

Table A1 in Appendix A contains threshold accuracy $A_{th}$, threshold values $V_{th}$, type, and change limits for all features. Values of threshold accuracy $A_{th}$ are sorted in descending order. Case distribution histograms for features with the highest threshold accuracy (LDL, HDL-C, Cholesterol, MCHC, Triglyceride, Amylase) are shown in Figure 6. An LDL level lower than 116.1 mg/dL, HDL-C level lower than 43.1 mg/dL, Cholesterol level lower than 206.3 mg/dL, Triglyceride level lower than 163.3 mg/dL, MCHC level higher than 31.3 g/dL, and Amylase level higher than 76.3 u/L mg/dL are critical levels for the detection of sick individuals. Considering any of these critical levels, the patients and healthy individuals could be detected with accuracy between $A_{th}$ = 85% and $A_{th}$ = 94%.

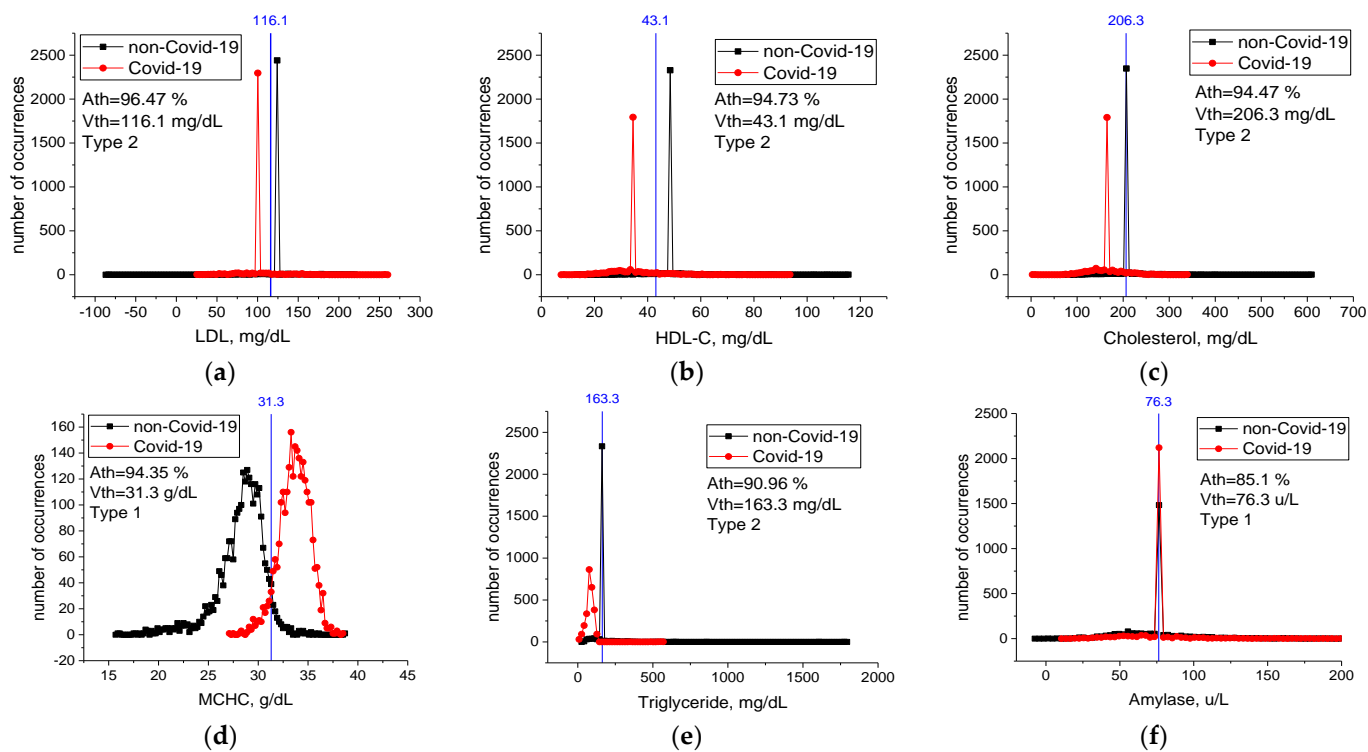

**Figure 6.** Case distribution histograms for LDL (**a**), HDL-C (**b**), Cholesterol (**c**), MCHC (**d**), Triglyceride (**e**), Amylase (**f**) from sick and healthy individuals and the threshold values $V_{th}$ of these features (blue line) in the diagnosis of the disease. Histogram bin sizes are listed in Table A1.

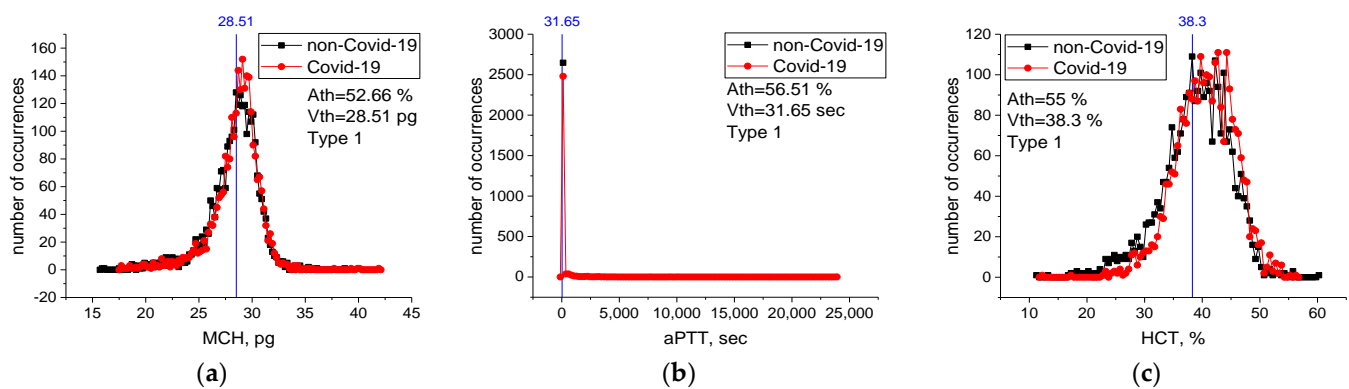

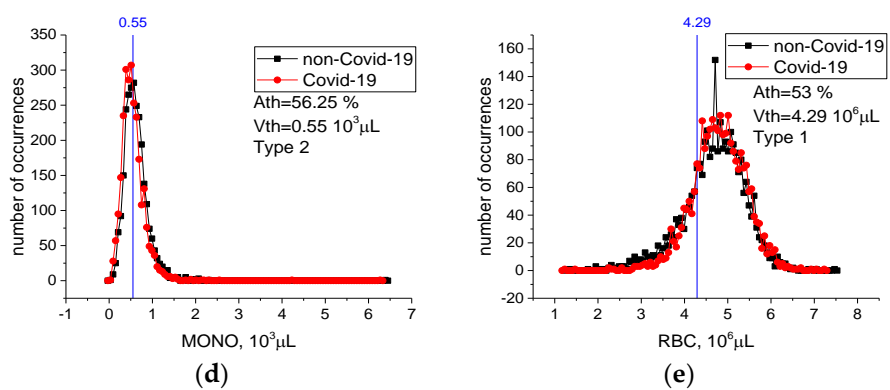

**Figure 7.** Case distribution histograms for MCH (a), aPTT (b), HCT (c), MONO (d), RBC (e) from sick and healthy individuals and the threshold values *Vth* of these features in the diagnosis of the disease. Histogram bin sizes are listed in table A1.

For features from Table 6 not included in Figure 6, case distribution histograms (MCH, aPTT, HCT, MONO, RBC) are demonstrated in Figure 7. The success of these features alone in detecting sick and healthy individuals was less than 60% (Figure 7). However, the combination of MCHC with MCH and the combination of MCHC with HDL-C in detecting sick and healthy individuals is higher than their individual performance (Table 7). Revealed high-level mutual information among these variables helps LogNNet to diagnose COVID-19. The combinations of MCH, aPTT, HCT, MONO, and RBC features are not effective in the diagnosis of the disease ($A_5(FS$ [10,17,19,22,25]), Table 7). We think that there is a low correlation between these features and COVID-19.

### 3.2. Dataset SARS-CoV-2-RBV2

LogNNet 51:50:20:2 architecture was used for the SARS-CoV-2-RBV2 dataset. The result of reservoir optimization obtained following the method from Section 2.3 with the number of epochs $Ep = 50$ led to the parameters of the congruential generator indicated in Table 4. Feature selection was carried out with the number of epochs $Ep = 150$. Prior to selection, feature strength corresponded to $dA_{51}(z)$ (Figure 3a). After feature selection, the redundant features are with numbers $z = 44, 45$ and $14$, and the $dA_{48}(z)$ graph is shown in Figure 3d.

The dependence of $A_{48}(FR$ [14,44,45]) on the number of epochs is shown in Figure 8, and the values of other metrics are shown in Table 8.

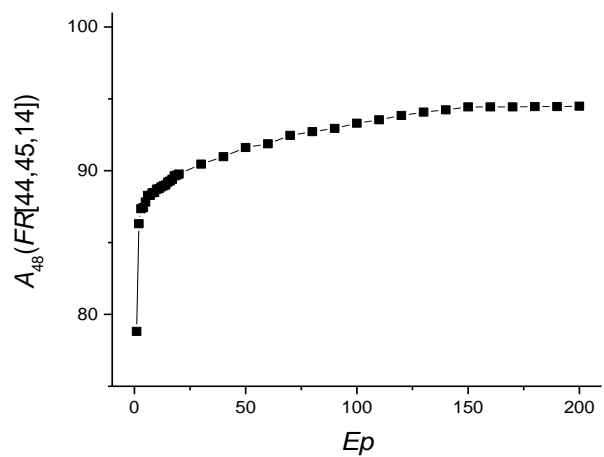

**Figure 8.** Dependence of $A_{48}(FR$ [14,44,45]) on the number of epochs $Ep$.

**Table 8.** Classification metrics depending on the number of training epochs *Ep*.

| Ep | $A_{48}$(*FR* [14,44,45]) | Precision "Non-ICU" | Precision "ICU" | Recall "Non-ICU" | Recall "ICU" | F1 "Non-ICU" | F2 "ICU" |
|---|---|---|---|---|---|---|---|
| 10 | 88.715 | 0.993 | 0.307 | 0.887 | 0.881 | 0.937 | 0.451 |
| 30 | 90.459 | 0.993 | 0.347 | 0.906 | 0.876 | 0.947 | 0.492 |
| 100 | 93.306 | 0.990 | 0.433 | 0.939 | 0.821 | 0.964 | 0.562 |
| 150 | 94.434 | 0.989 | 0.49 | 0.952 | 0.797 | 0.97 | 0.599 |
| 200 | 94.486 | 0.987 | 0.495 | 0.955 | 0.767 | 0.97 | 0.592 |

*Ep* = 150 is be taken as the optimal value of the number of epochs. The metrics for the "ICU" case are significantly worse than for the "non-ICU" case because of limited data for the "ICU" case. The most important RBVs in identifying severely and mildly infected COVID-19 patients are the features listed in Table 9. The most important of these are ESR and NEU.

**Table 9.** The 12 features found to be most important in detecting severely (ICU) and mildly (non-ICU) infected COVID-19 patients.

| Number | $dA_{48}$ | Features |
|---|---|---|
| 49 | 2.18 | ESR |
| 36 | 1.872 | NEU |
| 42 | 1.59 | CRP |
| 3 | 1.359 | Albumin |
| 39 | 1.154 | RBC |
| 12 | 0.974 | Chlorine |
| 40 | 0.872 | RDW |
| 4 | 0.795 | ALP |
| 21 | 0.795 | TP |
| 9 | 0.769 | Glucose |
| 35 | 0.744 | MPV |
| 29 | 0.718 | HGB |

ESR: erythrocyte sedimentation rate; NEU: neutrophil count; CRP: C-reactive protein; RBC: red blood cells; RDW: red cell distribution width; ALP: alkaline phosphatase; TP: total protein; MPV: mean platelet volume; HGB: hemoglobin.

The efficiency of LogNNet when using only the 12 features and their combinations to identify severely and mildly infected COVID-19 patients are shown in Table 10.

**Table 10.** LogNNet efficiency for various combinations of features.

| Combinations of Features | A | Precision "Non-ICU" | Precision "ICU" | Recall "Non-ICU" | Recall "ICU" | F1 "Non-ICU" | F1 "ICU" |
|---|---|---|---|---|---|---|---|
| $A_{48}$(*FR* [14,44,45]) | 94.434 | 0.989 | 0.49 | 0.952 | 0.797 | 0.97 | 0.599 |
| $A_{12}$(*FS* [3,4,9,12,21,29,35,36,39,40,42,49]) | 90.946 | 0.990 | 0.364 | 0.914 | 0.831 | 0.950 | 0.499 |
| $A_1$(*FS* [49]) | 59.598 | 0.950 | 0.059 | 0.605 | 0.418 | 0.694 | 0.097 |
| $A_1$(*FS* [39]) | 75.040 | 0.955 | 0.085 | 0.773 | 0.341 | 0.851 | 0.133 |
| $A_3$(*FS* [36,42,49]) | 82.712 | 0.989 | 0.210 | 0.827 | 0.826 | 0.900 | 0.334 |
| $A_7$(*FS* [3,12,36,39,40,42,49]) | 89.355 | 0.991 | 0.341 | 0.896 | 0.846 | 0.940 | 0.469 |

The recall value indicates what percentage of individuals diagnosed as mild or severe patients by the specialist could be recognized as mild or severe patients by our model. In other words, the recall value indicates the success of our model in distinguishing mild or severe patients. The precision value indicates the percentage of the individuals diagnosed as mild or severe patients by our model who were also defined as mild or severe patients by the specialist. In other words, the precision value shows the success of our model in diagnosing mild or severe patients.

The accuracy of the model run with 12 features to identify mildly and severely infected patients was close to the accuracy rate of the model run with 48 features ($A_{12}$~90.9 vs. $A_{48}$~94.94) (Table 10). The accuracy with the seven features model run was 89.3%, where the model success in diagnosing the mildly infected (precision value) was 99.1%, and success in recognizing mildly infected patients (recall value) was 89.6%. The metrics for the "ICU" case are significantly worse than for the "non-ICU" case. Here, our model decided in favor of the diagnosis of mildly infected (high precision for non-ICU, low precision for ICU) due to the sample number unbalance of our mildly infected and severely infected patients.

Threshold Accuracy on One Feature

Table A2 in Appendix A contains values of threshold accuracy $A_{th}$, threshold values $V_{th}$, as well as types and limits of change for all features. Rows in the table are sorted in descending order of threshold accuracy $A_{th}$. Case distribution histograms for features with the highest threshold accuracy (NEU, Albumin, WBC, CRP, Urea, Calcium) are shown in Figure 9.

Cases with an NEU level higher than $6.2 \times 10^3/\mu L$, WBC level higher than $7.93 \times 10^3/\mu L$, CRP level higher than 15 mg/dL, Urea level higher than 46.9 mg/dL, Albumin level lower than 32.2 g/L, and Calcium level lower than 8.5 mg/dL most likely require intensive care treatment (Figure 9). Considering any of these critical levels, patients requiring intensive care and patients not requiring intensive care could be correctly identified with the accuracy between $A_{th}$ = 72% and $A_{th}$ = 78%.

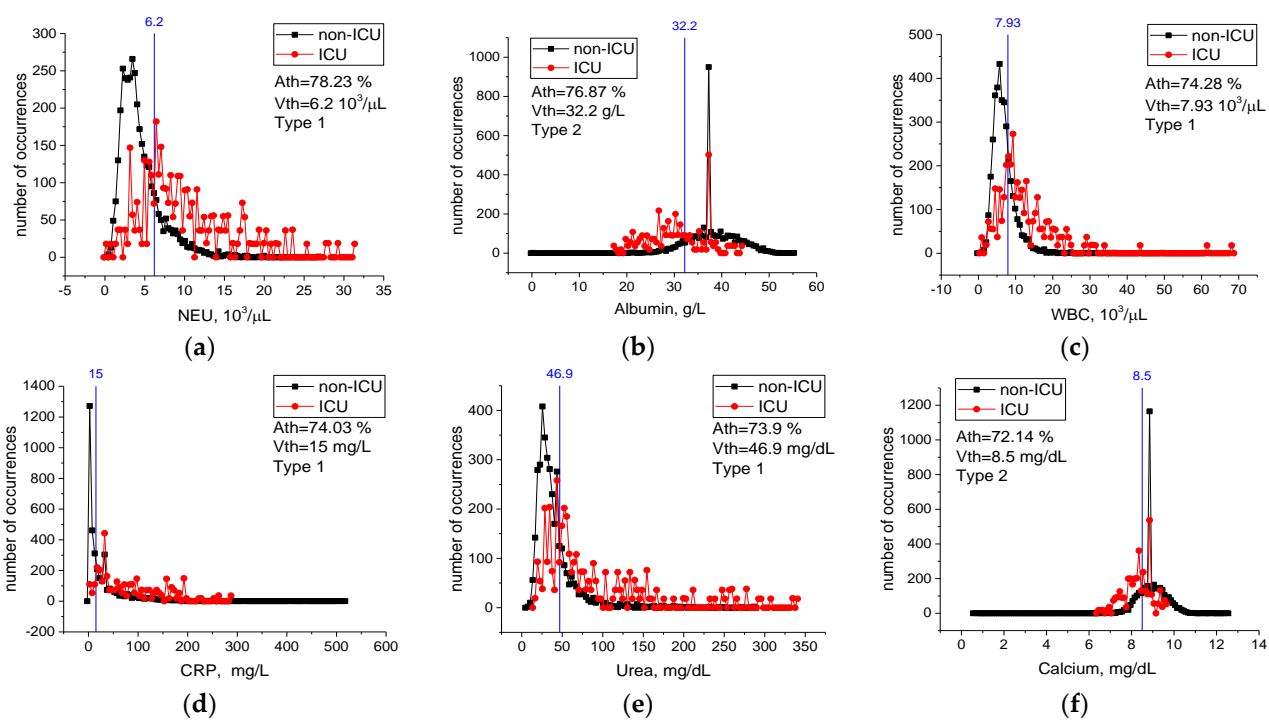

**Figure 9.** Case distribution histograms for NEU (**a**), Albumin (**b**), WBC (**c**), CRP (**d**), Urea (**e**), Calcium (**f**) from mildly and severely infected COVID-19 patients and the threshold values $V_{th}$ of these features (blue line) in the prognosis of the disease. Histogram bin sizes are listed in Table A2.

For features from Table 9 not included in Figure 9, case distribution histograms (ESR, RBC, Chlorine, RDW, ALP, TP, Glucose, MPV, HGB) are demonstrated in Figure 10. The success of these features alone in detecting mildly and severely infected patients varies between $Ath$ = 54.3% and $Ath$ = 71.5% (Figure 10). However, the performance of the combination of the ESR, NEU, and CRP features in detecting mild and severely infected patients was higher than their individual performance (Table 10). In addition, combinations of these properties with the Albumin, RBC, Chlorine, and RDW properties improved performance in detecting severely and mildly infected patients [$A_3$(FS [36,42,49] = 82.7% vs. $A_7$(FS [3,12,36,39,40,42,49] = 89.3% (Table 10). We think that there is a low level of correlation between the characteristics of ALP, TP, Glucose, MPV, and HGB and the severity of COVID-19 ($A_7$(FS [3,12,36,39,40,42,49])) = 89.4% vs. $A_{12}$(FS [3,4,9,12,21,29,35,36,39,40,42,49]) = 90.9% (Table 10). Therefore, the combination of the ESR, NEU, CRP, Albumin, RBC, Chlorine, and RDW blood values is an important source of variation in determining the severity of the disease, and high-level confidential information may be found among these variables. The combination of these features may have important effects in the prognosis of COVID-19 disease and in identifying patients in need of intensive care.

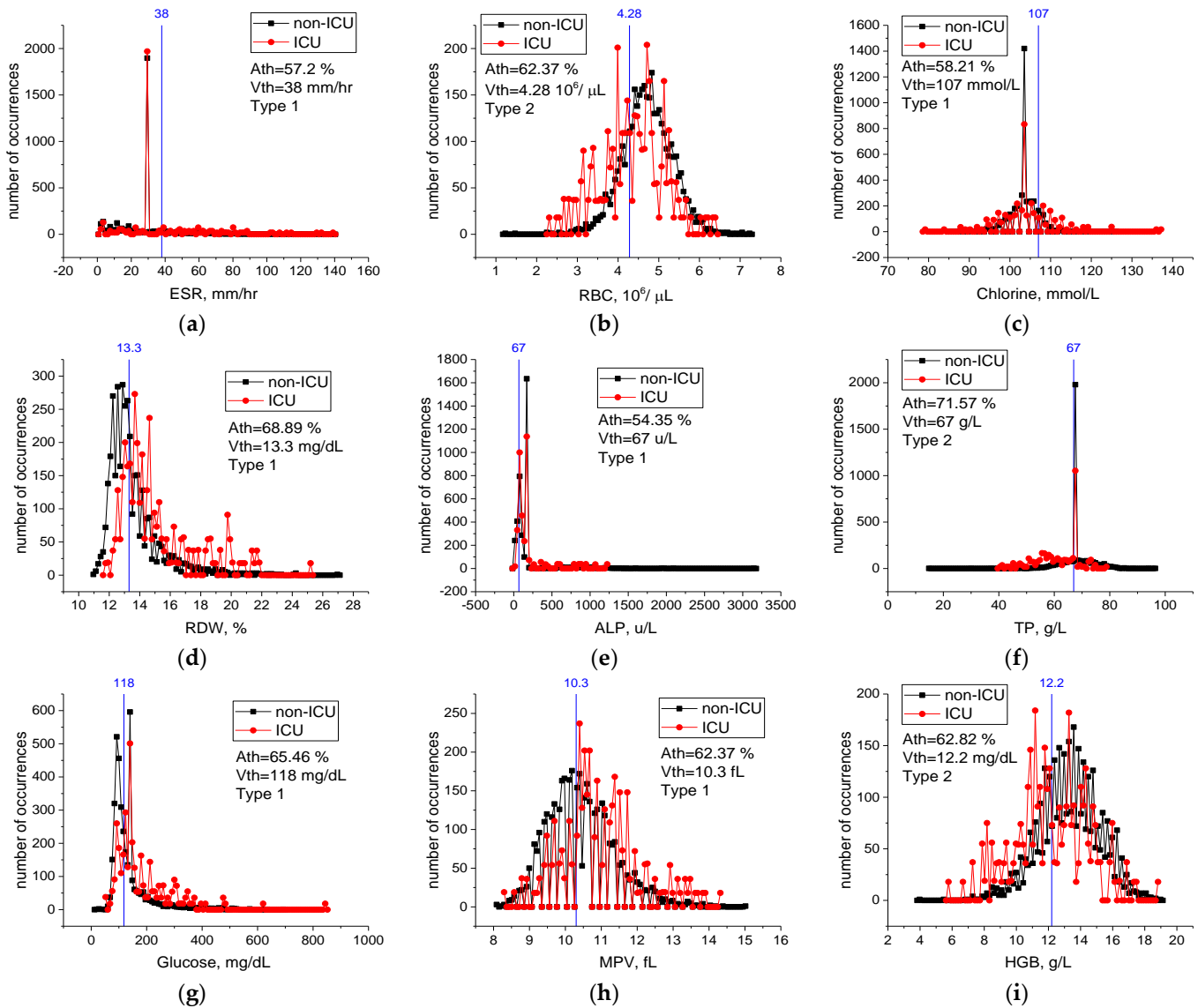

**Figure 10.** Case distribution histograms for ESR (**a**), RBC (**b**), Chlorine (**c**), RDW (**d**), ALP (**e**), TP(**f**), Glucose (**g**), MPV (**h**), HGB (**i**) from mildly and severely infected COVID-19 patients and the threshold values *Vth* of these features (blue line) in the prognosis of the disease. Histogram bin sizes are listed in Table A2.

## 4. Discussion

COVID-19 is a systemic multi-organ damage disease that causes severe acute respiratory syndrome, death, and continues to spread [3,47]. Despite the use of vaccines, the spread of the disease cannot be stopped, and important mutations have been detected in the structure of the virus [1]. It is likely that COVID-19 will continue to be present in our lives. Despite the large number of studies on COVID-19, some of these studies were contradictory and pathological aspects of the disease could not be fully determined [48]. Changes in many RBVs and hematological abnormalities were observed during the course of the disease [6,48]. The fact that most patients lost their lives in case of severe infection has led to a fight against the disease all over the world [10,49]. In addition, Brinati et al. [19] and Zhang et al. [49] pointed out that various complications may occur during the treatment process of COVID-19, and this makes it important to predict the prognosis of the disease in the early period. Similarly, Mertoğlu et al. [1] and Huyut and İlkbahar [3] stated that the early prediction of the diagnosis and prognosis of the disease are important in the first response to severely infected COVID-19 patients.

As with immunodiagnostic testing, RT-PCR testing may present difficulties in identifying true positive and negative individuals infected with COVID-19 [4,50]. Indeed, Teymouri et al. [50] and D'Cruz et al. [51] suggested that to increase the sensitivity of the RT-PCR test, the test should be repeated on multiple samples and the application methodology should be improved. However, these procedures represent a troublesome process for health personnel and patients. These difficulties in diagnosing COVID-19 have further increased the importance of RBVs methods [1,2]. In this context, it is possible to determine both the diagnosis and the prognosis of the disease with RBVs (biomarkers), which are easier to obtain, more economical, and faster to measure [1–6].

In an ML study for the diagnosis of COVID-19 based on RBVs, Brinati et al. [19] explained that AI models are based on clinical features and can be used for processes, such as disease diagnosis and prognosis. AI models that use the RBVs can be both an adjunct and an alternative method to rRT-PCR [20]. In addition, AI application results can provide information about the infection risk level and can be used in the rapid triage and quarantine of high-risk patients [20].

In this study, the most effective RBV biomarkers in the diagnosis and prognosis of COVID-19 were determined by a two-step feature selection procedure for use in peripheral IoT devices with low computing resources. Our LogNNet neural network model, fed with selected features, identified sick and healthy individuals, and especially mildly infected patients, with high accuracy.

In the first dataset used in this study, the RBVs of COVID-19 positive (*n* = 2648) patients and COVID-19 negative (*n* = 2648) individuals were recorded. In the second dataset, the RBVs of 3899 patients (*n* = 203 ICU and *n* = 3696 non-ICU) hospitalized with the diagnosis of COVID-19 were recorded. Hence, 51 features of all patients were identified (Tables 1 and 2). A two-stage feature selection procedure (see Section 2.5) was applied on the datasets and features were found for each dataset. The features selected for the first dataset were fed into the LogNNet neural network, and the accuracy of the method in the diagnosis of COVID-19 was calculated. Then, the selected features for the second dataset were fed into LogNNet neural network, and the performance of the method in identifying mildly and severely infected patients (determining the prognosis of the disease) was assessed.

Previous studies on the diagnosis and prognosis of COVID-19 have indicated the changes in most of the RBV parameters and biomarkers [1–3,5]. Mertoglu et al. [1] and Yang et al. [52] reported that the most effective RBV biomarkers in the diagnosis and prognosis of COVID-19 are CRP and LYM. However, other studies conducted for this purpose have reported blood values of CRP, procalcitonin, ferritin, ALT, aPTT, and ESR [3,4,6]. Banerjee et al. [8] used random forest, glmnet, generalized linear models, and ANN neural network models to determine the diagnosis of COVID-19 with 14 RBV values of 81 COVID-19 positive and 517 healthy individuals. Glmnet was found to be the most successful model in the diagnosis of the disease with 92% sensitivity and 91% accuracy [8]. Brinati et al. [19] used various ML methods with 13 RBV values for diagnosis of the disease (102 COVID-19 negative, 177 positive) and noted that the models with the highest accuracy were random forest (82%) and logistic regression (78%). Similarly, Cabitza et al. [20] used various ML models to rapidly detect COVID-19 using many RBV parameters and found the models with the highest accuracy were random forest (88%), support vector machine (SVM) (88%), and k-nearest neighbor (86%). Joshi et al. [22] developed a trained logistic regression model using some RBVs on a dataset of 380 cases, reporting good sensitivity (93%) but low specificity (43%). Yang et al. [21] applied various ML models on 27 RBV parameters of a large patient population of 3356 individuals (42% COVID-19 positive), and found the gradient boost tree model to be the most successful model in the diagnosis of the disease with 76%-sensitivity and 80%-specificity value. In a COVID-19 study using chest computed tomography (CT) data and RBV parameters, Mei et al. [23] showed a model combining CNN and multilayer sensor and found the success of the model in diagnosing the disease with 84% sensitivity and 83% specificity. Soares [24] proposed a model combining SVM, ensembling, and SMOTE Boost models to diagnose COVID-19 using 15 RBV parameters in a population of 599 individuals, and found the success of the model in diagnosing the disease with 86% specificity and 70% sensitivity. Running various ML models to diagnose COVID-19 with the RBV parameters, Soltan et al. [25] found the XGBoost method to be the most successful model with 85% sensitivity and 90% precision. Huyut [53] used 28 routine blood values with age on a variety of supervised ML models to detect a large population of severely and mildly infected COVID-19 patients. The models with the highest AUC in identifying mildly infected patients were local weighted-learning (0.95%), Kstar (0.91%), Naïve bayes (0.85%), and K nearest neighbor (0.75%).

This study identified the seven most important biomarkers in the diagnosis of COVID-19 (Table 6). Among these features, the most important biomarkers were MCHC, MCH, and aPTT. The overall accuracy rate of the LogNNet model, which was run with seven features, was $A_7(FS\ [10,17,19,20,22,25,36])$ ~99.3%, and the precision rate of patient identification was 99.6%. In addition, the different combinations of features that are important in the diagnosis of patients were examined. The overall accuracy of the LogNNet model run only with MCHC and MCH features was $A_2(FS\ [19,20])$ ~99.1% and the precision rate of patient identification was 99.4%. The overall accuracy rate of our model using only the MCHC feature was 94.2%, while the overall accuracy rate of the model using only the HDL-C feature was 94.4%. According to the calculated critical levels of the main features, such as LDL, HDL-C, Cholesterol, Triglyceride, MCHC, and Amylase (Figure 6), the health and sickness status of individuals could be determined accurately. The fact that the performance of the combination of MCHC and MCH and the combination of MCHC and HDL-C in the detection of sick and healthy individuals was higher than the individual performances suggested that there is a high level of confidential information between these blood feature combinations and COVID-19. This information was revealed by the LogNNet neural network method. These combinations of features can be used by LognNNet in diagnosis of COVID-19 disease with high results.

Studies indicate that the ALT, AST, LDH, direct bilirubin, and aPTT RBVs are increased in severe COVID-19 patients, while the hemoglobin values are decreased significantly compared to mildly infected patients [6,23,54]. However, in other studies, the LYM, NEU, WBC, MCH, MPV, and RDW hematological RBVs were higher in severe COVID-19 patients, when compared to mildly infected patients [1–3,6]. Mousavi et al. [16], Zhang et al. [54], and Zheng et al. [55] determined that patients with severe COVID-19 had lower EOS, MONO, RBC, hematocrit, hemoglobin, and MCHC hematological values, when compared to mild patients. Huyut et al. [6], in a study of patients who died from COVID-19, showed that the ESR, INR, PT, CRP, D-dimer, and ferritin biomarkers are the most important biomarkers to detect the mortality of the disease. Luo et al. [56] proposed a multi-criteria decision making (MCDM) algorithm combining ideal the solution similarity sequencing technique (TOPSIS) and naive Bayes (NB) as a feature selection procedure to predict the severity of COVID-19 from initial RBV values. With the MCDM model, the WBC, LYM, NEU values, and age were the most effective features in determining the severity of the disease with 82% accuracy obtained by ROC analysis [56]. Similarly, Ma et al. [57] and Lai et al. [58] noted that the high WBC and NEU values are important manifestations of bacterial infection and indicate a serious disease state that complicates the clinical situation. Numerous studies have shown that other proinflammatory marker levels, including CRP, ferritin, and IL-6, are associated with worse outcomes [59–61]. Cheng et al. [62] reported that high levels of inflammatory markers, such as ESR, CRP, and procalcitonin, may indicate hyperinflammatory reactions in COVID-19 patients. Cavalcante-Silva et al. [63] stated that the neutrophil count was increased in severe COVID-19 patients and the neutrophils are the main effector cells in the development of COVID-19. The different neutrophil mechanisms, e.g., neutrophil enzymes and cytokines, are potential targets for treating particularly severe cases of COVID-19 [63].

This study identifies the twelve most important biomarkers to determine the prognosis of COVID-19 (detecting severely and mildly infected patients) (Table 9). The most important of them are ESR, NEU, CRP, albumin, and RBC biomarkers. The overall accuracy of the LogNNet model, which was run with twelve features, was 90.9%, the success rate in diagnosing mildly infected patients (precision rate) was 99.0%, and the success rate in diagnosing severely infected patients (precision rate) was 36.6% (Table 10). However, the success of the LogNNet model, which was run with twelve features, in distinguishing mild and severe patients according to their real conditions (recall value), was 91.4% and 83.1%, respectively (Table 10).

The calculated critical levels of NEU, WBC, CRP, Urea, Albumin, and Calcium features are important levels in determining the severity of infection of the patients (Figure 9). Moreover, the performance of the combination of the ESR, NEU, CRP, Albumin, RBC, Chlorine, and RDW features in detecting infected patients being higher than their individual performance indicates a high level of confidential information about COVID-19 among these blood features. This information was revealed by the LogNNet neural network. The combinations of features can be used as important biomarkers in the prognosis of the COVID-19 disease and in identifying patients in need of intensive care.

Our model decided in favor of the diagnosis of mildly infected patients (high precision for non-ICU, low precision for ICU) because of the unbalanced sample size of mildly infected and severely infected patients. However, our model showed a high recall value in identifying mildly and severely infected patients. The model run with only three features showed an average of 82.6% agreement with the expert opinion in distinguishing mildly or severely infected patients (Table 10). However, severe patient diagnosis of our model showed low agreement with expert opinion (low precision "ICU") (Table 10), and the success of our model in diagnosing severe patients is low. As a result, the LogNNet model, which is run with the features in Table 10, can be used safely with high sensitivity (recall) to confirm the expert opinion in recognizing mild and severely infected patients. In addition, our model can be an alternative tool for diagnosing mildly infected patients using the features in Table 10. Furthermore, the success of the LogNNet model using few features in distinguishing mild and severe patients and diagnosing mildly infected patients is high.

Other studies [19,64,65] confirming the association of RBV features with COVID-19 highlight the importance of the clinical research direction that our model takes. The poor performance of our model in diagnosing severe patients (low precision for the ICU) is an expected situation. Several studies have stated that severe COVID-19 patients experienced more changes in the RBV values than mildly infected patients, and that various complications could occur during the severe disease process [1–3,6]. There are many factors affecting the intensive care need of an individual with COVID-19 and difficulties in determining this process with only RBV values [1–6]. However, there are few studies on determining the severity of infection in patients with COVID-19 based on the RBV values alone.

Cabitza et al. [20], Soltan et al. [25], and Rabanser et al. [66] stated that the reported performance values are good enough, especially in terms of screening, considering the economic benefits and rapid results of the developed artificial intelligence models. Moreover, Brinati et al. [19] suggested the necessity of conducting studies on the predictability of arterial blood gas tests in addition to routine blood values for the diagnosis of COVID-19. In this context, we plan our next studies as follows. The first phase is to identify the diagnosis and prognosis of COVID-19 with LogNNet model using the arterial blood gases. The next phase is to determine the mortality of COVID-19 with the LogNNet model using the RBV values.

Velichko [43] reported a method for the estimation of the occupied RAM in the implementation of the LogNNet on Arduino microcontrollers. The LogNNet 51:50:20:2 model, discussed above, takes about 13.7 kB of RAM. As the matrix *W* occupies ~10.4 kB, this memory can be freed due to RAM saving algorithm, and the algorithm will use ~3.3 kB. Therefore, the model can be placed on microcontrollers with a RAM size of 16 kB, e.g., Arduino Nano.

With recent advancements in information and communication technologies due to the adoption of IoT technology, smart health monitoring and support systems have a higher development and acceptability margin to improve wellness [67,68]. The integration of medical technologies into IoT is called the Internet of Medical Things (IoMT) [69].

In this context, the availability of low-cost, single-chip microcontrollers and advances in wireless communication technology have encouraged researchers to design low-cost embedded systems for healthcare monitoring applications [67]. Doctors can use patients' data to remotely monitor their physiological health status and diagnose their disorders [68]. In a study designed for mobile health applications, Hu et al. [70] used various graphical biosensors to monitor conditions, such as heart attack, brain problems, and high blood pressure (seizures, mental disorder, etc.). In a study for a similar purpose, Vizbaras et al. [71] reported that the stretching and bending vibrations of various chemical bonds are molecule-specific. Therefore, certain infrared spectral ranges are of particular interest in biomedical sensing. In addition, this approach can be used to selectively detect important biomolecules, such as glucose, lactate, urea, ammonia, serum albumin, and so on. Clifton et al. [72] demonstrated the use of wearable sensors for routine healthcare in their study of the large-scale clinical adoption of "intelligent" predictive monitoring systems.

Mobile sensors for the measurement of routine blood parameters to be used in the real-time detection of various diseases are being developed rapidly with the advancements of technology [73–76]. The RBV values can be measured using a low-cost, mobile microscope, an ocular camera, and a smartphone [73]. Chan et al. [74] determined PT and INR blood values by monitoring the micro-mechanical movements of a copper particle with a proof-of-concept using the vibration motor and camera in smartphones. Farooqi et al. [75] followed the diabetic patients with telemonitoring and Bluetooth-enabled self-monitoring devices and produced new solutions for the glycemic control of the patients. Zhang et al.[76] determined various biochemical parameters by electrochemical controls.

In the feature, the data can be obtained in real time and used to provide immediate medical advice before the health problems of the patients occur and progress. The technique presented in this study can be used to create mobile health monitoring systems.

The output of the LogNNet model can be used in different scenarios. The presented feature selection method can be used in conjunction with molecular testing to obtain high sensitivity and certainty regarding suspected cases. In this way, more positive patients can be identified, isolated, and treated in a timely manner. Likewise, the outputs of our model can be used while the results of other tests are awaited. The results of this study demonstrated that the LogNNet neural network model can be used with high productivity for clinical decision support systems and mobile diagnostics.

Various independent biomarkers used in the study need to be tested in the diagnosis and prognosis of other infectious diseases. The low number of ICU patient groups compared to the non-ICU group was one of the limitations of this study.

## 5. Conclusions

Determining the mild or severe infection status of COVID-19 patients using various diagnostic tests and imaging results can be costly, time consuming, and is subject to different complications during the process. In this case, the patient's health may be at higher risk and health services may face tragic situations under intense pressure. This study provides a fast, reliable, and economic alternative mobile tool for the diagnosis and prognosis of COVID-19 based on the RBV values measured only at the time of admission to the hospital.

In this study, the most effective RBVs in the diagnosis and prognosis of COVID-19 were determined using a feature selection method for the LogNNet reservoir neural network. The most important RBVs in the diagnosis of the disease were MCHC, MCH, and aPTT. The most important RBVs in the prognosis of the disease were ESR, NEU, CRP, albumin, and RBC. The LogNNet deep neural network model accurately and precisely detected almost all COVID-19 patients using only a few RBV features.

The health and sickness status of individuals could be determined largely accurately using threshold levels of the LDL, HDL-C, Cholesterol, Triglyceride, MCHC, and Amylase features. In addition, the LogNNet neural network revealed that the performance of the combination of MCHC and MCH and the combination of MCHC and HDL-C in the detection of sick and healthy individuals was higher than the individual performances of these features.

Threshold levels of the NEU, WBC, CRP, Urea, Albumin, and Calcium main properties were found to be significant in the detection of severely and mildly infected patients. As revealed by the LogNNet network, the combination of ESR, NEU, CRP, Albumin, RBC, Chlorine, and RDW features is an important source of variation in the prognosis of COVID-19. We propose to use this combination of the features with LogNNet as important biomarkers in the prognosis of the disease and in identifying patients in need of intensive care.

The results of this study can be effectively used in medical peripheral devices of the IoT (IoTM) with low RAM resources, including clinical decision support systems, remote internet medicine, and telemedicine.

**Supplementary Materials:** The following supporting information can be downloaded at: www.mdpi.com/xxx/s1, SARS-CoV-2-RBV (1-2)_Dataset.zip, Readme.pdf. Kindly cite our paper when you wish to use this dataset.

**Author Contributions:** Conceptualization, M.T.H. and A.V.; methodology, M.T.H. and A.V.; software, A.V.; validation, M.T.H. and A.V.; formal analysis, M.T.H.; investigation, A.V.; resources, M.T.H.; data curation, M.T.H.; writing—original draft preparation, M.T.H., and A.V.; writing—review and editing, M.T.H. and A.V.; visualization, M.T.H. and A.V.; supervision, M.T.H.; project administration, M.T.H.; funding acquisition, A.V. All authors have read and agreed to the published version of the manuscript.

**Funding:** This research was supported by the Russian Science Foundation (grant no. 22-11-00055, https://rscf.ru/en/project/22-11-00055/).

**Institutional Review Board Statement:** The dataset used in this study was collected in order to be used in various studies in the estimation of the diagnosis, prognosis and mortality of COVID-19. The necessary permissions for the collected dataset were given by the Ministry of Health of the Republic of Turkey and the Ethics Committee of Erzincan Binali Yıldırım University. This study was conducted in accordance with the 1989 Declaration of Helsinki. Erzincan Binali Yıldırım University Human Research Health and Sports Sciences Ethics Committee Decision Number: 2021/02-07.

**Informed Consent Statement:** In this study, a dataset including only routine blood values, RT-PCR results (positive or negative) and treatment units of the patients was downloaded retrospectively from the information system of our hospital in digital environment. A new sample was not taken from the patients. There is no information in the dataset that includes identifying characteristics of individuals. It was stated that routine blood values would only be used in academic studies, and written consent was obtained from the institutions for this. In addition, therefore, written informed consent was not administered for every patient.

**Data Availability Statement:** The data used in this study can be shared with the parties, provided that the article is cited.

**Acknowledgments:** We thank the method of Erzincan Mengücek Gazi Training and Research Hospital for their support in reaching the material used in this study. Special thanks to the editors of the journal and to the anonymous reviewers for their constructive criticism and improvement suggestions.

**Conflicts of Interest:** The authors declare no conflict of interest.

## Appendix A

**Table A1.** Threshold method parameters for SARS-CoV-2-RBV1 dataset and histogram bin sizes for Figures 6 and 7.

| № | Feature | $A_{th}$, % | $V_{th}$ | Units | Type | Min | Max | Bin size |
|---|---|---|---|---|---|---|---|---|
| 43 | LDL | 96.47 | 116.14 | mg/dL | 2 | −83 | 258 | 3.4 |
| 36 | HDL-C | 94.73 | 43.09 | mg/dL | 2 | 8 | 115 | 1 |
| 39 | Cholesterol | 94.47 | 206.33 | mg/dL | 2 | 5 | 606 | 6 |
| 20 | MCHC | 94.35 | 31.31 | g/dL | 1 | 15.9 | 38.6 | 0.2 |
| 48 | Triglyceride | 90.96 | 163.35 | mg/dL | 2 | 34 | 1782 | 17 |
| 31 | Amylase | 85.1 | 76.35 | u/L | 1 | 0 | 1193 | 3 |
| 51 | UA | 81.12 | 5.39 | mg/dL | 1 | 0 | 14.3 | |
| 47 | TP | 79.68 | 68.05 | g/L | 2 | 15 | 96 | |
| 32 | CK-MB | 78.91 | 19.87 | u/L | 2 | 0 | 685.5 | |
| 42 | LDH | 74.98 | 258.40 | u/L | 1 | 0 | 2749 | |
| 29 | Albumin | 74.91 | 39.61 | g/L | 2 | 0 | 55.87 | |
| 37 | Calcium | 74.21 | 9.01 | mg/dL | 2 | 0 | 12.55 | |
| 30 | ALP | 74.13 | 154.35 | u/L | 1 | 0 | 3150 | |
| 38 | Chlorine | 72.62 | 103.47 | mmol/L | 2 | 79 | 345 | |
| 34 | GGT | 71.6 | 35.51 | u/L | 1 | 0 | 2732 | |
| 1 | CRP | 70.54 | 4.29 | mg/L | 1 | 1 | 1650 | |
| 41 | CK | 70.47 | 111.96 | u/L | 2 | 0 | 4665 | |
| 45 | Sodium | 69.24 | 139.02 | mmol/L | 1 | 108 | 175 | |
| 3 | Ferritin | 68.75 | 49.69 | µg/L | 1 | 0.2 | 1650 | |
| 46 | T-Bil | 68.52 | 0.58 | mg/dL | 2 | −0.35 | 20.95 | |
| 33 | D-Bil | 66.09 | 0.16 | mg/dL | 2 | −0.06 | 20 | |
| 11 | LYM | 66.01 | 1.50 | $10^3/\mu L$ | 2 | 0.08 | 715 | |
| 40 | Creatinine | 64.03 | 1.01 | mg/dL | 1 | 0 | 202 | |
| 7 | PCT | 63.22 | 0.12 | ng/mL | 1 | 0.12 | 1500 | |
| 4 | Fibrinogen | 63.18 | 307.94 | mg/dL | 2 | 10.9 | 668.07 | |
| 35 | Glucose | 62.42 | 122.05 | mg/dL | 1 | 11 | 846 | |
| 49 | eGFR | 61.48 | 87.22 | no unit | 2 | 3.483 | 561.746 | |
| 27 | ALT | 61.35 | 29.54 | u/L | 1 | 0 | 2110 | |
| 28 | AST | 60.65 | 32.19 | u/L | 1 | 0 | 2927 | |

| № | Feature | Ath, % | Vth | Units | Type | Min | Max | Bin Size |
|---|---|---|---|---|---|---|---|---|
| 2 | D-Dimer | 60.37 | 385.41 | μg/L | 2 | 1.06 | 9610 | |
| 50 | Urea | 58.19 | 40.99 | mg/dL | 1 | 0 | 427 | |
| 14 | WBC | 58.08 | 5.71 | 10³/μL | 2 | 0.4 | 127 | |
| 13 | PLT | 57.46 | 200.26 | 10³/μL | 2 | 9 | 768 | |
| 8 | ESR | 57.38 | 14.07 | mm/hr | 1 | 2 | 124 | |
| 16 | EOS | 56.4 | 0 | 10³/μL | 1 | 0 | 4.41 | |
| 21 | MCV | 56.25 | 84.03 | fL | 1 | 56.7 | 122.1 | |
| 22 | MONO | 56.25 | 0.54 | 10³/μL | 2 | 0.03 | 6.4 | 0.06 |
| 44 | Potassium | 55.63 | 4.36 | mmol/L | 1 | 0 | 59 | |
| 26 | RDW | 55.49 | 13.21 | % | 2 | 0 | 30.8 | |
| 15 | BASO | 55.04 | 0.029 | 10³/μL | 2 | 0 | 0.38 | |
| 17 | HCT | 55 | 38.33 | % | 1 | 11.4 | 60.1 | 60 |
| 10 | aPTT | 56.51 | 31.06 | Sec | 1 | 12 | 23,843.7 | 238 |
| 12 | NEU | 54.8 | 2.60 | 10³/μL | 2 | 0.49 | 66.43 | |
| 18 | HGB | 54.12 | 12.31 | g/L | 1 | 3.7 | 19 | |
| 5 | INR | 53.15 | 0.735 | no unit | 2 | 0.12 | 88 | |
| 25 | RBC | 53 | 4.29 | 10⁶/μL | 1 | 1.24 | 7.48 | 0.06 |
| 19 | MCH | 52.66 | 28.51 | pg | 1 | 15.9 | 41.9 | 0.2 |
| 24 | PDW | 51.93 | 11.89 | fL | 1 | 0 | 25.3 | |
| 23 | MPV | 51.79 | 9.81 | fL | 1 | 0 | 15 | |
| 6 | PT | 51.79 | 13.09 | Sec | 1 | 2 | 181 | |
| 9 | Troponin | 50.19 | 25 | ng/L | 1 | 0.01 | 25,000 | |

**Table A2.** Threshold method parameters for SARS-CoV-2-RBV2 dataset and histogram bin sizes for Figures 9 and 10.

| № | Feature | Ath, % | Vth | Units | Type | Min | Max | Bin Size |
|---|---|---|---|---|---|---|---|---|
| 36 | NEU | 78.23 | 6.20 | 10³/μL | 1 | 0.1 | 31.26 | 0.3 |
| 3 | Albumin | 76.87 | 32.20 | g/L | 2 | 0.08 | 55 | 0.5 |
| 41 | WBC | 74.28 | 7.93 | 10³/μL | 1 | 0.4 | 68.3 | 0.6 |
| 42 | CRP | 74.03 | 15.051 | mg/L | 1 | 0.15 | 514 | 5 |
| 24 | Urea | 73.92 | 46.95 | mg/dL | 1 | 6 | 339 | 3 |
| 11 | Calcium | 72.14 | 8.50 | mg/dL | 2 | 0.6 | 12.43 | 0.1 |
| 21 | TP | 71.57 | 67.00 | g/L | 2 | 15 | 96 | 0.8 |
| 30 | LYM | 71.48 | 1.02 | 10³/μL | 2 | 0.08 | 58.87 | |
| 40 | RDW | 68.89 | 13.30 | % | 1 | 11 | 27 | 0.16 |
| 48 | PCT | 67.85 | 0.151 | ng/mL | 1 | 0.052 | 100 | |
| 2 | AST | 66.39 | 44.92 | u/L | 1 | 4 | 2927 | |
| 16 | LDH | 66.11 | 267.37 | u/L | 1 | 20 | 1547 | |
| 9 | Glucose | 65.46 | 118.13 | mg/dL | 1 | 17 | 846 | 8 |
| 7 | D-Bil | 65.04 | 0.209 | mg/dL | 1 | 0.01 | 20 | |
| 44 | Ferritin | 64.17 | 238.116 | μg/L | 1 | 2.4 | 2000 | |
| 15 | CK | 63.66 | 99.92 | u/L | 1 | 2 | 4665 | |
| 43 | D-Dimer | 63.61 | 1074 | μg/L | 1 | 1.06 | 37,000 | |
| 29 | HGB | 62.82 | 12.20 | g/L | 2 | 4 | 19 | 0.15 |
| 47 | PT | 62.78 | 14.30 | Sec | 1 | 9.4 | 129 | |
| 23 | eGFR | 62.55 | 80.47 | no unit | 2 | 4.724 | 561.746 | |
| 35 | MPV | 62.37 | 10.30 | fL | 1 | 8.1 | 15 | 0.07 |
| 39 | RBC | 62.37 | 4.28 | 10⁶/μL | 2 | 1.24 | 7.22 | 0.06 |
| 50 | Troponin | 61.86 | 10.19 | ng/L | 1 | 1 | 4600 | |
| 20 | T-Bil | 61.81 | 0.58 | mg/dL | 1 | 0.01 | 29 | |
| 8 | GGT | 61.41 | 57.36 | u/L | 1 | 1 | 1085 | |
| 19 | Sodium | 61.01 | 145 | mmol/L | 1 | 112 | 175 | |
| 37 | PDW | 60.86 | 11.51 | fL | 1 | 7.6 | 25.3 | |
| 32 | MCHC | 60.72 | 32.11 | g/dL | 2 | 3.6 | 39.2 | |
| 28 | HCT | 59.71 | 36.63 | % | 2 | 12 | 56.3 | |
| 1 | ALT | 59.02 | 39.80 | u/L | 1 | 0.7 | 1349 | |
| 33 | MCV | 58.79 | 85.93 | fL | 1 | 55.8 | 117.8 | |
| 6 | CK-MB | 58.72 | 19.38 | u/L | 1 | 1 | 575.4 | |
| 14 | Creatinine | 58.39 | 1.26 | mg/dL | 1 | 0.46 | 202 | |
| 12 | Chlorine | 58.21 | 107 | mmol/L | 1 | 79 | 137 | 0.58 |
| 45 | Fibrinogen | 57.22 | 334 | mg/dL | 1 | 70.56 | 681.88 | |
| 49 | ESR | 57.2 | 38.03 | mm/hr | 1 | 2 | 139 | 1.37 |
| 5 | Amylase | 56.46 | 75.7 | $10^3/\mu L$ | 2 | 11 | 874 | |
| 46 | INR | 56.38 | 1.42 | no unit | 1 | 0.77 | 110 | |
| 51 | aPTT | 56.33 | 36.12 | Sec | 2 | 12 | 414 | |
| 25 | UA | 55.92 | 5.412 | mg/dL | 1 | 0.9 | 15 | |
| 38 | PLT | 55.61 | 160 | % | 2 | 5 | 1199 | |
| 34 | MONO | 55.22 | 0.474 | sec | 2 | 0.03 | 6.29 | |
| 18 | Potassium | 54.99 | 3.815 | mmol/L | 2 | 2.4 | 59 | |
| 27 | EOS | 54.72 | 0.111 | $10^3/Ml$ | 2 | 0.01 | 4.41 | |
| 4 | ALP | 54.35 | 63.98 | u/L | 1 | 1 | 3150 | 31 |
| 22 | Triglyceride | 53.27 | 141.6 | $10^6/\mu L$ | 1 | 32 | 1402 | |
| 31 | MCH | 53.11 | 28.22 | pg | 2 | 15.6 | 41.9 | |
| 13 | Cholesterol | 53.11 | 170 | mg/dL | 2 | 5 | 354 | |
| 10 | HDL-C | 53.02 | 34.69 | mg/dL | 2 | 8 | 93 | |
| 26 | BASO | 52.75 | 0.01 | $10^3/\mu L$ | 1 | 0.01 | 0.38 | |
| 17 | LDL | 51.26 | 115.1 | mg/dL | 1 | 15 | 258 | |

**References**

1. Mertoglu, C.; Huyut, M.; Olmez, H.; Tosun, M.; Kantarci, M.; Coban, T. COVID-19 is more dangerous for older people and its severity is increasing: A case-control study. *Med. Gas Res.* **2022**, *12*, 51–54. https://doi.org/10.4103/2045-9912.325992.
2. Mertoglu, C.; Huyut, M.T.; Arslan, Y.; Ceylan, Y.; Coban, T.A. How do routine laboratory tests change in coronavirus disease 2019? *Scand. J. Clin. Lab. Investig.* **2021**, *81*, 24–33. https://doi.org/10.1080/00365513.2020.1855470.
3. Huyut, M.T.; İlkbahar, F. The effectiveness of blood routine parameters and some biomarkers as a potential diagnostic tool in the diagnosis and prognosis of Covid-19 disease. *Int. Immunopharmacol.* **2021**, *98*, 107838. https://doi.org/10.1016/j.intimp.2021.107838.
4. HUYUT, M.T.; HUYUT, Z. Forecasting of Oxidant/Antioxidant levels of COVID-19 patients by using Expert models with biomarkers used in the Diagnosis/Prognosis of COVID-19. *Int. Immunopharmacol.* **2021**, *100*, 108127. https://doi.org/10.1016/j.intimp.2021.108127.
5. Huyut, M.; Üstündağ, H. Prediction of diagnosis and prognosis of COVID-19 disease by blood gas parameters using decision trees machine learning model: A retrospective observational study. *Med. Gas Res.* **2022**, *12*, 60–66. https://doi.org/10.4103/2045-9912.326002.
6. Tahir Huyut, M.; Huyut, Z.; İlkbahar, F.; Mertoğlu, C. What is the impact and efficacy of routine immunological, biochemical and hematological biomarkers as predictors of COVID-19 mortality? *Int. Immunopharmacol.* **2022**, *105*, 108542. https://doi.org/10.1016/j.intimp.2022.108542.
7. Guan, W.; Ni, Z.; Hu, Y.; Liang, W.; Ou, C.; He, J.; Liu, L.; Shan, H.; Lei, C.; Hui, D.S.C.; et al. Clinical Characteristics of Coronavirus Disease 2019 in China. *N. Engl. J. Med.* **2020**, *382*, 1708–1720. https://doi.org/10.1056/NEJMOA2002032.
8. Banerjee, A.; Ray, S.; Vorselaars, B.; Kitson, J.; Mamalakis, M.; Weeks, S.; Baker, M.; Mackenzie, L.S. Use of Machine Learning and Artificial Intelligence to predict SARS-CoV-2 infection from Full Blood Counts in a population. *Int. Immunopharmacol.* **2020**, *86*, 106705. https://doi.org/10.1016/J.INTIMP.2020.106705.
9. Huyut, M.T.; Soygüder, S. The Multi-Relationship Structure between Some Symptoms and Features Seen during the New Coronavirus 19 Infection and the Levels of Anxiety and Depression post-Covid. *East. J. Med.* **2022**, *27(1)*, 1-10. https://doi.org/10.5505/ejm.2022.35336.
10. Amgalan, A.; Othman, M. Hemostatic laboratory derangements in COVID-19 with a focus on platelet count. *Platelets* **2020**, *31*, 740–745. https://doi.org/10.1080/09537104.2020.1768523.

11. Li, X.; Wang, L.; Yan, S.; Yang, F.; Xiang, L.; Zhu, J.; Shen, B.; Gong, Z. Clinical characteristics of 25 death cases with COVID-19: A retrospective review of medical records in a single medical center, Wuhan, China. *Int. J. Infect. Dis.* **2020**, *94*, 128–132. https://doi.org/10.1016/j.ijid.2020.03.053.
12. Kukar, M.; Gunčar, G.; Vovko, T.; Podnar, S.; Černelč, P.; Brvar, M.; Zalaznik, M.; Notar, M.; Moškon, S.; Notar, M. COVID-19 diagnosis by routine blood tests using machine learning. *Sci. Rep.* **2021**, *11*, 10738. https://doi.org/10.1038/s41598-021-90265-9.
13. Jiang, S.Q.; Huang, Q.F.; Xie, W.M.; Lv, C.; Quan, X.Q. The association between severe COVID-19 and low platelet count: Evidence from 31 observational studies involving 7613 participants. *Br. J. Haematol.* **2020**, *190*, e29–e33. https://doi.org/10.1111/bjh.16817.
14. Zheng, Y.; Zhang, Y.; Chi, H.; Chen, S.; Peng, M.; Luo, L.; Chen, L.; Li, J.; Shen, B.; Wang, D. The hemocyte counts as a potential biomarker for predicting disease progression in COVID-19: A retrospective study. *Clin. Chem. Lab. Med.* **2020**, *58*, 1106–1115. https://doi.org/10.1515/cclm-2020-0377.
15. Lippi, G.; Plebani, M.; Henry, B.M. Thrombocytopenia is associated with severe coronavirus disease 2019 (COVID-19) infections: A meta-analysis. *Clin. Chim. Acta* **2020**, *506*, 145–148. https://doi.org/10.1016/j.cca.2020.03.022.
16. Mousavi, S.A.; Rad, S.; Rostami, T.; Rostami, M.; Mousavi, S.A.; Mirhoseini, S.A.; Kiumarsi, A. Hematologic predictors of mortality in hospitalized patients with COVID-19: A comparative study. *Hematology* **2020**, *25*, 383–388. https://doi.org/10.1080/16078454.2020.1833435.
17. Beck, B.R.; Shin, B.; Choi, Y.; Park, S.; Kang, K. Predicting commercially available antiviral drugs that may act on the novel coronavirus (SARS-CoV-2) through a drug-target interaction deep learning model. *Comput. Struct. Biotechnol. J.* **2020**, *18*, 784–790. https://doi.org/10.1016/j.csbj.2020.03.025.
18. Xu, X.; Jiang, X.; Ma, C.; Du, P.; Li, X.; Lv, S.; Yu, L.; Ni, Q.; Chen, Y.; Su, J.; et al. A Deep Learning System to Screen Novel Coronavirus Disease 2019 Pneumonia. *Engineering* **2020**, *6*, 1122–1129. https://doi.org/10.1016/j.eng.2020.04.010.
19. Brinati, D.; Campagner, A.; Ferrari, D.; Locatelli, M.; Banfi, G.; Cabitza, F. Detection of COVID-19 Infection from Routine Blood Exams with Machine Learning: A Feasibility Study. *J. Med. Syst.* **2020**, *44*, 135. https://doi.org/10.1007/s10916-020-01597-4.
20. Cabitza, F.; Campagner, A.; Ferrari, D.; Di Resta, C.; Ceriotti, D.; Sabetta, E.; Colombini, A.; De Vecchi, E.; Banfi, G.; Locatelli, M.; et al. Development, evaluation, and validation of machine learning models for COVID-19 detection based on routine blood tests. *Clin. Chem. Lab. Med.* **2021**, *59*, 421–431. https://doi.org/10.1515/cclm-2020-1294.
21. Yang, H.S.; Hou, Y.; Vasovic, L.V.; Steel, P.A.D.; Chadburn, A.; Racine-Brzostek, S.E.; Velu, P.; Cushing, M.M.; Loda, M.; Kaushal, R.; et al. Routine Laboratory Blood Tests Predict SARS-CoV-2 Infection Using Machine Learning. *Clin. Chem.* **2020**, *66*, 1396–1404. https://doi.org/10.1093/clinchem/hvaa200.
22. Joshi, R.P.; Pejaver, V.; Hammarlund, N.E.; Sung, H.; Kyu, S.; Lee, H.; Scott, G.; Gombar, S.; Shah, N.; Shen, S.; et al. Short communication A predictive tool for identification of SARS-CoV-2 PCR-negative emergency department patients using routine test results. *J. Clin. Virol.* **2020**, *129*, 104502. https://doi.org/10.1016/j.jcv.2020.104502.
23. Mei, X.; Lee, H.C.; Diao, K.Y.; Huang, M.; Lin, B.; Liu, C.; Xie, Z.; Ma, Y.; Robson, P.M.; Chung, M.; et al. Artificial intelligence–enabled rapid diagnosis of patients with COVID-19. *Nat. Med.* **2020**, *26*, 1224–1228. https://doi.org/10.1038/s41591-020-0931-3.
24. Soares, F. A novel specific artificial intelligence-based method to identify COVID-19 cases using simple blood exams. *medRxiv* **2020**. https://doi.org/10.1101/2020.04.10.20061036.
25. Soltan, A.A.; Kouchaki, S.; Zhu, T.; Kiyasseh, D.; Taylor, T.; Hussain, Z.B.; Peto, T.; Brent, A.J.; Eyre, D.W.; Clifton, D. Artificial intelligence driven assessment of routinely collected healthcare data is an effective screening test for COVID-19 in patients presenting to hospital. *medRxiv* **2020**.
26. Remeseiro, B.; Bolon-Canedo, V. A review of feature selection methods in medical applications. *Comput. Biol. Med.* **2019**, *112*, 103375. https://doi.org/10.1016/j.compbiomed.2019.103375.
27. Bikku, T. Multi-layered deep learning perceptron approach for health risk prediction. *J. Big Data* **2020**, *7*, 50. https://doi.org/10.1186/s40537-020-00316-7.
28. Battineni, G.; Chintalapudi, N.; Amenta, F. Machine learning in medicine: Performance calculation of dementia prediction by support vector machines (SVM). *Inform. Med. Unlocked* **2019**, *16*, 100200. https://doi.org/10.1016/j.imu.2019.100200.
29. Xing, W.; Bei, Y. Medical Health Big Data Classification Based on KNN Classification Algorithm. *IEEE Access* **2020**, *8*, 28808–28819. https://doi.org/10.1109/ACCESS.2019.2955754.
30. Hoodbhoy, Z.; Noman, M.; Shafique, A.; Nasim, A.; Chowdhury, D.; Hasan, B. Use of machine learning algorithms for prediction of fetal risk using cardiotocographic data. *Int. J. Appl. Basic Med. Res.* **2019**, *9*, 226. https://doi.org/10.4103/ijabmr.ijabmr_370_18.
31. Alam, M.Z.; Rahman, M.S.; Rahman, M.S. A Random Forest based predictor for medical data classification using feature ranking. *Inform. Med. Unlocked* **2019**, *15*, 100180. https://doi.org/10.1016/j.imu.2019.100180.
32. Schober, P.; Vetter, T.R. Logistic Regression in Medical Research. *Anesth. Analg.* **2021**, *132*, 365–366. https://doi.org/10.1213/ANE.0000000000005247.
33. Podgorelec, V.; Kokol, P.; Stiglic, B.; Rozman, I. Decision trees: An overview and their use in medicine. *J. Med. Syst.* **2002**, *26*, 445–463.
34. Guyon, I.; Gunn, S.; Nikravesh, M.; Zadeh, L.A. *Feature Extraction: Foundations and Applications*; Studies in Fuzziness and Soft Computing; Springer: Berlin/Heidelberg, Germany, 2008; ISBN 9783540354888.
35. Hall, M.A. Correlation-based Feature Selection for Machine Learning. PhD Thesis, Department of Computer Science, The University of Waikato, Hamilton, NewZealand, *April* 1999, 51-69.

36. Dash, M.; Liu, H. Consistency-based search in feature selection. *Artif. Intell.* **2003**, *151*, 155–176. https://doi.org/10.1016/S0004-3702(03)00079-1.
37. Zhao, Z.; Liu, H. Searching for interacting features. *20 th International Joint Conference on Artificial Intelligence, Hyderabad, India, 6-12 January, Code 97873, 2007* pp.1156–1161.
38. Hall, M.A.; Smith, L.A. Practical feature subset selection for machine learning. In *Computer Science '98, Proceedings of the 21st Australasian Computer Science Conference ACSC'98, Perth, Australia, 4–6 February 1998*; Volume 20, pp. 181–191.
39. Kononenko, I. Estimating attributes: Analysis and extensions of RELIEF. In *European Conference on Machine Learning*; Springer: Berlin/Heidelberg, Germany, 1994; Volume 784, pp. 171–182. https://doi.org/10.1007/3-540-57868-4_57.
40. Le Thi, H.A.; Nguyen, V.V.; Ouchani, S. Gene selection for cancer classification using DCA. In *International Conference on Advanced Data Mining and Applications*; Springer: Berlin/Heidelberg, Germany, 2008; Volume 5139, pp. 62–72. https://doi.org/10.1007/978-3-540-88192-6_8.
41. Tibshirani, R. Regression Shrinkage and Selection via the Lasso. *J. R. Stat. Soc. Ser. B* **1996**, *58*, 267–288.
42. Velichko, A. Neural network for low-memory IoT devices and MNIST image recognition using kernels based on logistic map. *Electronics* **2020**, *9*, 1432. https://doi.org/10.3390/electronics9091432.
43. Velichko, A. A method for medical data analysis using the lognnet for clinical decision support systems and edge computing in healthcare. *Sensors* **2021**, *21*, 6209. https://doi.org/10.3390/s21186209.
44. Velichko, A.; Heidari, H. A Method for Estimating the Entropy of Time Series Using Artificial Neural Networks. *Entropy* **2021**, *23*, 1432. https://doi.org/10.3390/e23111432.
45. Izotov, Y.A.; Velichko, A.A.; Boriskov, P.P. Method for fast classification of MNIST digits on Arduino UNO board using LogNNet and linear congruential generator. *J. Phys. Conf. Ser.* **2021**, *2094*, 32055. https://doi.org/10.1088/1742-6596/2094/3/032055.
46. Heidari, H.; Velichko, A. An improved LogNNet classifier for IoT application. *J. Phys. Conf. Ser.* **2021**, *2094*, 032015.
47. Mattiuzzi, C.; Lippi, G. Which lessons shall we learn from the 2019 novel coronavirus outbreak? *Ann. Transl. Med.* **2020**, *8*, 48–48. https://doi.org/10.21037/atm.2020.02.06.
48. Kim, S.; Kim, D.-M.; Lee, B. Insufficient Sensitivity of RNA Dependent RNA Polymerase Gene of SARS-CoV-2 Viral Genome as Confirmatory Test using Korean COVID-19 Cases. *Preprints* **2020**, 1–4. https://doi.org/10.20944/preprints202002.0424.v1.
49. Zhang, J.J.; Cao, Y. yuan; Tan, G.; Dong, X.; Wang, B. chen; Lin, J.; Yan, Y. qin; Liu, G. hui; Akdis, M.; Akdis, C.A.; et al. Clinical, radiological, and laboratory characteristics and risk factors for severity and mortality of 289 hospitalized COVID-19 patients. *Allergy Eur. J. Allergy Clin. Immunol.* **2021**, *76*, 533–550. https://doi.org/10.1111/all.14496.
50. Teymouri, M.; Mollazadeh, S.; Mortazavi, H.; Naderi Ghale-noie, Z.; Keyvani, V.; Aghababaei, F.; Hamblin, M.R.; Abbaszadeh-Goudarzi, G.; Pourghadamyari, H.; Hashemian, S.M.R.; et al. Recent advances and challenges of RT-PCR tests for the diagnosis of COVID-19. *Pathol. Res. Pract.* **2021**, *221*, 153443. https://doi.org/10.1016/j.prp.2021.153443.
51. D'Cruz, R.J.; Currier, A.W.; Sampson, V.B. Laboratory Testing Methods for Novel Severe Acute Respiratory Syndrome-Coronavirus-2 (SARS-CoV-2). *Front. Cell Dev. Biol.* **2020**, *8*, 468. https://doi.org/10.3389/fcell.2020.00468.
52. Yang, A.P.; Liu, J.P.; Tao, W. qiang; Li, H. ming The diagnostic and predictive role of NLR, d-NLR and PLR in COVID-19 patients. *Int. Immunopharmacol.* **2020**, *84*, 106504. https://doi.org/10.1016/j.intimp.2020.106504.
53. Huyut, M.T. Automatic Detection of Severely and Mildly Infected COVID-19 Patients with Supervised Machine Learning Models. *IRBM* **2022**, *1*, 1–12. https://doi.org/10.1016/j.irbm.2022.05.006.
54. Zhang, C.; Shi, L.; Wang, F.S. Liver injury in COVID-19: Management and challenges. *Lancet Gastroenterol. Hepatol.* **2020**, *5*, 428–430. https://doi.org/10.1016/S2468-1253(20)30057-1.
55. Zheng, M.; Gao, Y.; Wang, G.; Song, G.; Liu, S.; Sun, D.; Xu, Y.; Tian, Z. Functional exhaustion of antiviral lymphocytes in COVID-19 patients. *Cell. Mol. Immunol.* **2020**, *17*, 533–535. https://doi.org/10.1038/s41423-020-0402-2.
56. Luo, J.; Zhou, L.; Feng, Y.; Li, B.; Guo, S. The selection of indicators from initial blood routine test results to improve the accuracy of early prediction of COVID-19 severity. *PLoS ONE* **2021**, *16*, e0253329. https://doi.org/10.1371/journal.pone.0253329.
57. Ma, Y.; Hou, L.; Yang, X.; Huang, Z.; Yang, X.; Zhao, N.; He, M.; Shi, Y.; Kang, Y.; Yue, J.; et al. The association between frailty and severe disease among COVID-19 patients aged over 60 years in China: A prospective cohort study. *BMC Med.* **2020**, *18*, 274. https://doi.org/10.1186/s12916-020-01761-0.
58. Lai, C.C.; Shih, T.P.; Ko, W.C.; Tang, H.J.; Hsueh, P.R. Severe acute respiratory syndrome coronavirus 2 (SARS-CoV-2) and coronavirus disease-2019 (COVID-19): The epidemic and the challenges. *Int. J. Antimicrob. Agents* **2020**, *55*, 105924. https://doi.org/10.1016/j.ijantimicag.2020.105924.
59. Feld, J.; Tremblay, D.; Thibaud, S.; Kessler, A.; Naymagon, L. Ferritin levels in patients with COVID-19: A poor predictor of mortality and hemophagocytic lymphohistiocytosis. *Int. J. Lab. Hematol.* **2020**, *42*, 773–779. https://doi.org/10.1111/ijlh.13309.
60. Zhou, F.; Yu, T.; Du, R.; Fan, G.; Liu, Y.; Liu, Z.; Xiang, J.; Wang, Y.; Song, B.; Gu, X.; et al. Clinical course and risk factors for mortality of adult inpatients with COVID-19 in Wuhan, China: A retrospective cohort study. *Lancet* **2020**, *395*, 1054–1062. https://doi.org/10.1016/S0140-6736(20)30566-3.
61. Chen, G.; Wu, D.; Guo, W.; Cao, Y.; Huang, D.; Wang, H.; Wang, T.; Zhang, X.; Chen, H.; Yu, H.; et al. Clinical and immunological features of severe and moderate coronavirus disease 2019. *J. Clin. Investig.* **2020**, *130*, 2620–2629. https://doi.org/10.1172/JCI137244.
62. Cheng, L.; Li, H.; Li, L.; Liu, C.; Yan, S.; Chen, H.; Li, Y. Ferritin in the coronavirus disease 2019 (COVID-19): A systematic review and meta-analysis. *J. Clin. Lab. Anal.* **2020**, *34*, 1–18. https://doi.org/10.1002/jcla.23618.

63. Cavalcante-Silva, L.H.A.; Carvalho, D.C.M.; Lima, É.D.A.; Galvão, J.G.; da Silva, J.S.d.F.; de Sales-Neto, J.M.; Rodrigues-Mascarenhas, S. Neutrophils and COVID-19: The road so far. *Int. Immunopharmacol.* **2021**, *90*, 107233. https://doi.org/10.1016/j.intimp.2020.107233.
64. Pan, F.; Ye, T.; Sun, P.; Gui, S.; Liang, B.; Li, L.; Zheng, D.; Wang, J.; Hesketh, R.L.; Yang, L.; et al. Time Course of Lung Changes on Chest CT During Recovery From 2019 Novel Coronavirus (COVID-19) Pneumonia. *Radiology* **2020**, *295*, 200370. https://doi.org/10.1148/radiol.2020200370.
65. Zhao, D.; Yao, F.; Wang, L.; Zheng, L.; Gao, Y.; Ye, J.; Guo, F.; Zhao, H.; Gao, R. A Comparative Study on the Clinical Features of Coronavirus 2019 (COVID-19) Pneumonia with Other Pneumonias. *Clin. Infect. Dis.* **2020**, *71*, 756–761.
66. Rabanser, S.; Günnemann, S.; Lipton, Z.C. Failing loudly: An empirical study of methods for detecting dataset shift. *Adv. Neural Inf. Process. Syst.* **2019**, *32*. https://doi.org/10.48550/arXiv.1810.11953.
67. Al-Aubidy, K.M.; Derbas, A.M.; Al-Mutairi, A.W. Real-time patient health monitoring and alarming using wireless-sensor-network. In Proceedings of the 2016 13th International Multi-Conference on Systems, Signals & Devices (SSD), Leipzig, Germany, 21–24 March 2016; pp. 416–423. https://doi.org/10.1109/SSD.2016.7473672.
68. Taiwo, O.; Ezugwu, A.E. Smart healthcare support for remote patient monitoring during Covid-19 quarantine. *Inform. Med. Unlocked* **2020**, *20*, 100428. https://doi.org/10.1016/j.imu.2020.100428.
69. Lamonaca, F.; Balestrieri, E.; Tudosa, I.; Picariello, F.; Carnì, D.L.; Scuro, C.; Bonavolontà, F.; Spagnuolo, V.; Grimaldi, G.; Colaprico, A. An Overview on Internet of Medical Things in Blood Pressure Monitoring. In Proceedings of the 2019 IEEE International Symposium on Medical Measurements and Applications (MeMeA), Istanbul, Turkey, 26–28 June 2019; pp. 1–6.
70. Hu, F.; Xiao, Y.; Hao, Q. Congestion-aware, loss-resilient bio-monitoring sensor networking for mobile health applications. *IEEE J. Sel. Areas Commun.* **2009**, *27*, 450–465. https://doi.org/10.1109/JSAC.2009.090509.
71. Vizbaras, A.; Simonyte, I.; Droz, S.; Torcheboeuf, N.; Miasojedovas, A.; Trinkunas, A.; Buciunas, T.; Dambrauskas, Z.; Gulbinas, A.; Boiko, D.L.; et al. GaSb Swept-Wavelength Lasers for Biomedical Sensing Applications. *IEEE J. Sel. Top. Quantum Electron.* **2019**, *25*, 1–12. https://doi.org/10.1109/JSTQE.2019.2915967.
72. Clifton, L.; Clifton, D.A.; Pimentel, M.A.F.; Watkinson, P.J.; Tarassenko, L. Predictive monitoring of mobile patients by combining clinical observations with data from wearable sensors. *IEEE J. Biomed. Health Inform.* **2014**, *18*, 722–730. https://doi.org/10.1109/JBHI.2013.2293059.
73. Pfeil, J.; Nechyporenko, A.; Frohme, M.; Hufert, F.T.; Schulze, K. Examination of blood samples using deep learning and mobile microscopy. *BMC Bioinform.* **2022**, *23*, 1–14. https://doi.org/10.1186/s12859-022-04602-4.
74. Chan, J.; Michaelsen, K.; Estergreen, J.K.; Sabath, D.E.; Gollakota, S. Micro-mechanical blood clot testing using smartphones. *Nat. Commun.* **2022**, *13*, 1–12. https://doi.org/10.1038/s41467-022-28499-y.
75. Farooqi, M.H.; Abdelmannan, D.K.; Mubarak, M.; Abdalla, M.; Hamed, A.; Xavier, M.; Joyce, T.; Cadiz, S.; Nawaz, F.A. The Impact of Telemonitoring on Improving Glycemic and Metabolic Control in Previously Lost-to-Follow-Up Patients with Type 2 Diabetes Mellitus: A Single-Center Interventional Study in the United Arab Emirates. *Int. J. Clin. Pract.* **2022**, *2022*, 6286574.
76. Zhang, Y.; Zhang, Y.; Li, H.; Cao, Y.; Han, S.; Zhang, K.; He, W. Covalent Biosensing Polymer Chain Reaction Enabling Periphery Blood Testing to Predict Tumor Invasiveness with a Platelet Procancerous Protein. *Anal. Chem.* **2022**, *94*, 1983–1989. https://doi.org/10.1021/acs.analchem.1c03349.